\documentclass[10pt,twocolumn,letterpaper]{article}

\usepackage{cvpr}              
\usepackage[table]{xcolor} 
\usepackage{multirow}
\usepackage{makecell} 

\definecolor{cvprblue}{rgb}{0.21,0.49,0.74}
\usepackage[pagebackref,breaklinks,colorlinks,allcolors=cvprblue]{hyperref}

\def\paperID{***} 
\def\confName{CVPR}
\def\confYear{2026}

\title{Chain of World: World Model Thinking in Latent Motion}

\author{
    Fuxiang Yang\textsuperscript{1,2}\thanks{Work done during an internship at Li Auto.}\quad
    Donglin Di\textsuperscript{2}\quad
    Lulu Tang\textsuperscript{3,6}\quad
    Xuancheng Zhang\textsuperscript{2}\quad
    Lei Fan\textsuperscript{4}\\
    Hao Li\textsuperscript{2}\quad
    Wei Chen\textsuperscript{2}\quad
    Tonghua Su\textsuperscript{1,5}\thanks{Corresponding author.}\quad
    Baorui Ma\textsuperscript{2}\thanks{Project leader and corresponding author.}
    \vspace{0.2cm}\\
    \textsuperscript{1}Harbin Institute of Technology \quad
    \textsuperscript{2}Li Auto \quad
    \textsuperscript{3}Beijing Academy of Artificial Intelligence (BAAI) \\
    \textsuperscript{4}University of New South Wales \quad
    \textsuperscript{5}Chongqing Research Institute of HIT \quad
    \textsuperscript{6}Peking University
    \vspace{0.1cm}\\
    {\tt\small hityangfx@foxmail.com, donglin.ddl@gmail.com, lulutang\_@outlook.com} \\
    {\tt\small xczhang.thu@gmail.com, lei.fan1@unsw.edu.au, \{lihao43, chenwei10\}@lixiang.com} \\
    {\tt\small thsu@hit.edu.cn, mabaorui2014@gmail.com}
}

\begin{document}
\maketitle
\begin{abstract}
Vision-Language-Action (VLA) models are a promising path toward embodied intelligence, yet they often overlook the predictive and temporal-causal structure underlying visual dynamics.
World-model VLAs address this by predicting future frames, but waste capacity reconstructing redundant backgrounds.
Latent-action VLAs encode frame-to-frame transitions compactly, but lack temporally continuous dynamic modeling and world knowledge.
To overcome these limitations, we introduce CoWVLA (Chain-of-World VLA), a new ``Chain of World'' paradigm that unifies world-model temporal reasoning with a disentangled latent motion representation. 
First, a pretrained video VAE serves as a latent motion extractor, explicitly factorizing video segments into structure and motion latents. 
Then, during pre-training, the VLA learns from an instruction and an initial frame to infer a continuous latent motion chain and predict the segment's terminal frame. 
Finally, during co-fine-tuning, this latent dynamic is aligned with discrete action prediction by jointly modeling sparse keyframes and action sequences in a unified autoregressive decoder.
This design preserves the world-model benefits of temporal reasoning and world knowledge while retaining the compactness and interpretability of latent actions, enabling efficient visuomotor learning.
Extensive experiments on robotic simulation benchmarks show that CoWVLA outperforms existing world-model and latent-action approaches and achieves moderate computational efficiency, highlighting its potential as a more effective VLA pretraining paradigm.
The project website can be found at https://fx-hit.github.io/cowvla-io.
\end{abstract}    
\section{Introduction}
\label{sec:intro}

\begin{figure}[t]
  \centering   
\includegraphics[width=\linewidth]{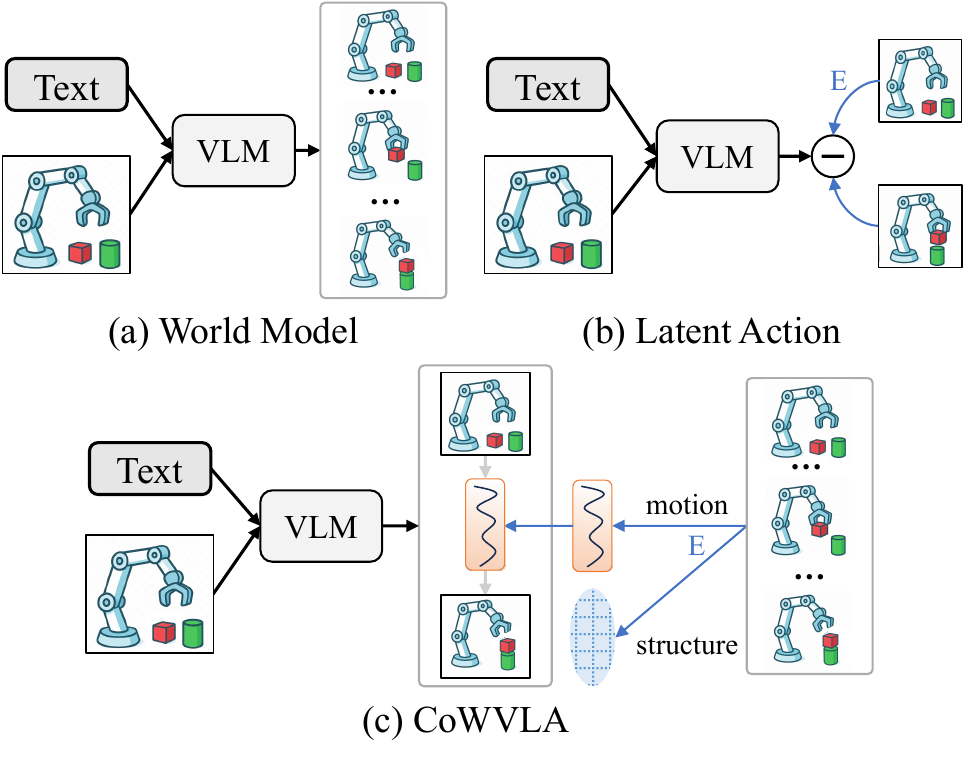}
   \vspace{-3ex}
   \caption{
   \textbf{Comparison of VLA pretraining strategies.}
   (a)~\textit{World Model}: It predicts future visual frames, leading to redundant background reconstruction.
   (b)~\textit{Latent Action}: It learns the frame-to-frame transition using a visual encoder $E$, but lacks temporally continuous reasoning.
   (c)~\textit{CoWVLA}: Our method first uses a video encoder $E$ to decompose each video segment into motion and structure latents, and then trains the VLM to infer latent motion and predict the terminal frame of the segment given the instruction and the initial frame.
   }
   \label{fig:intro}
   \vspace{-2ex}
\end{figure}

Embodied intelligence aims to build agents that can perceive, understand, and act in the physical world.
Vision-Language-Action (VLA) models represent a significant step toward this goal, unifying multimodal perception and motor control into end-to-end transformers~\cite{zitkovich2023rt2, kim2024openvla, black2024pi_0, o2024openx}.
While effective at mapping visual observations and language instructions directly to actions for many tasks, standard VLAs lack the future prediction capabilities that humans possess, which has spurred interest in enriching them with predictive world models~\cite{assran2025vjepa,bruce2024genie}.

A prominent approach integrates world models into VLAs by predicting future visual frames to explicitly model environmental dynamics, as illustrated in Figure~\ref{fig:intro} (a).
Methods such as WorldVLA~\cite{cen2025worldvla}, UniVLA~\cite{wang2025unified}, and FlowVLA~\cite{zhong2025flowvla} typically built on large-scale autoregressive transformers, learn to anticipate future states and thus benefit action policy learning.
While effective, this paradigm has fundamental limitations.
It requires modeling entire visual frames containing substantial static and redundant background pixels, leading to near-trivial pixel replication rather than focusing on meaningful motion and dynamic change.
Furthermore, quantizing images~\cite{esser2021vqgan} into discrete tokens results in excessively long sequences and severe training inefficiency when multiple frames are used.

From a cognitive standpoint, such frame prediction is misaligned with how humans model the world: we reason about motion and interactions rather than rebuilding every pixel in memory.
This observation raises an important question: \textit{can we build a more compact, abstract, and dynamic form of world modeling?}
The latent action paradigm~\cite{ye2025lap,chen2025moto,bu2025tla,chen2025villa} offers compelling inspiration.
As shown in Figure~\ref{fig:intro} (b), it encodes frame-to-frame transitions as latent actions, which serve as abstract motion carriers for world modeling, enabling large-scale pretraining using the pseudo-action labels built from videos.

However, we identify two critical limitations in the current latent action paradigm compared to world models.
First, world models perform temporally continuous dynamic modeling, whereas existing latent actions often focus only on the change between two frames~\cite{ye2025lap,chen2025moto,bu2025tla}.
Second, world models, through future frame prediction, learn generalizable knowledge for task execution and common sense about the world.
In contrast, latent actions only encode ``how to move'', but lack an understanding of what is moving, where the motion happens, or how the scene should evolve after the motion.

To address these limitations, we propose Chain-of-World VLA (CoWVLA), which establishes a new paradigm that unifies the advantages of both approaches, as shown in Figure~\ref{fig:intro} (c).
Our key insight is that effective world modeling requires both the compactness of motion representations and the temporal continuity and world knowledge of frame prediction.
We argue that it is possible to extract continuous and compact motion representations from video clips, suggesting the need for a model capable of decoupling the content structure and motion in videos.
Such motion representations serve as carriers for perceiving essential dynamic changes and further enable the model to reason about keyframes after temporal evolution, thereby preserving crucial visual landmarks.

Specifically, our approach employs a pretrained video VAE as a latent motion extractor, which explicitly disentangles each video segment into structure and motion representations, providing compact and interpretable supervision for downstream visuomotor learning.
We then train a unified VLA decoder through two stages.
During the pre-training stage, the model learns to infer latent dynamics and predict the terminal frame of a video segment given the instruction and initial frame, thereby establishing a dynamics-aware world prior in the latent motion space.
During the subsequent co-fine-tuning stage, this prior is further aligned with discrete action prediction by jointly modeling sparse keyframes and action sequences in a unified autoregressive manner.
This design combines the interpretability and compactness of latent motion with the temporal reasoning and world knowledge of world models, achieving efficient and robust visuomotor learning without reconstructing redundant intermediate frames.

In summary, our contributions are as follows: 
\begin{itemize}
    \item We present CoWVLA, establishing the ``Chain-of-World'' paradigm that unifies world modeling and latent action learning through continuous latent-motion sequences and terminal keyframe prediction.
    \item We introduce a structure-motion disentangled latent prior that yields interpretable, continuous, and effective dynamic representations. 
    \item We conduct extensive experiments demonstrating that CoWVLA achieves state-of-the-art performance across multiple benchmarks, surpassing existing world-model and latent-action approaches.
\end{itemize}

\section{Related Work}
\label{sec:related_work}

\textbf{Vision-Language-Action Models.}
Deep learning has been widely applied in various industrial scenarios, such as visual anomaly detection~\cite{fan2025salvaging,fan2025manta}.
Recent vision-language-action (VLA) models have rapidly advanced toward directly generating actions from visual and language inputs within a unified framework~\cite{zitkovich2023rt2, kim2024openvla,o2024openx,kim2025openvla_oft,qu2025spatialvla,pertsch2025fast,intelligence2025pi_0_5,black2024pi_0,gao2025vla_os,sun2025llapa,wang2026hmvla}. 
RT-2~\cite{zitkovich2023rt2} pioneered this direction by treating robotic control as a sequence modeling problem, fine-tuning a pretrained vision-language model on robotic data to output discretized action tokens. 
This approach was scaled up by RT-X~\cite{o2024openx}, which demonstrated the benefits of joint training across diverse robot platforms and tasks. OpenVLA~\cite{kim2024openvla,kim2025openvla_oft} further democratized this effort with an open-source implementation.
FAST~\cite{pertsch2025fast} introduced a unified frequency-domain formulation for discretizing actions, enhancing temporal correlation in discrete control.
Meanwhile, another line of research explores continuous trajectory generation~\cite{chi2023diffusionpolicy,black2024pi_0,li2024cogact,hou2025dita}.
They leverage diffusion or flow-matching models to generate continuous, high-frequency action sequences.
However, most existing methods primarily focus on action space modeling, with limited capability to capture how the environment evolves.

\begin{figure*}[t]
  \centering   
\includegraphics[width=1\linewidth]{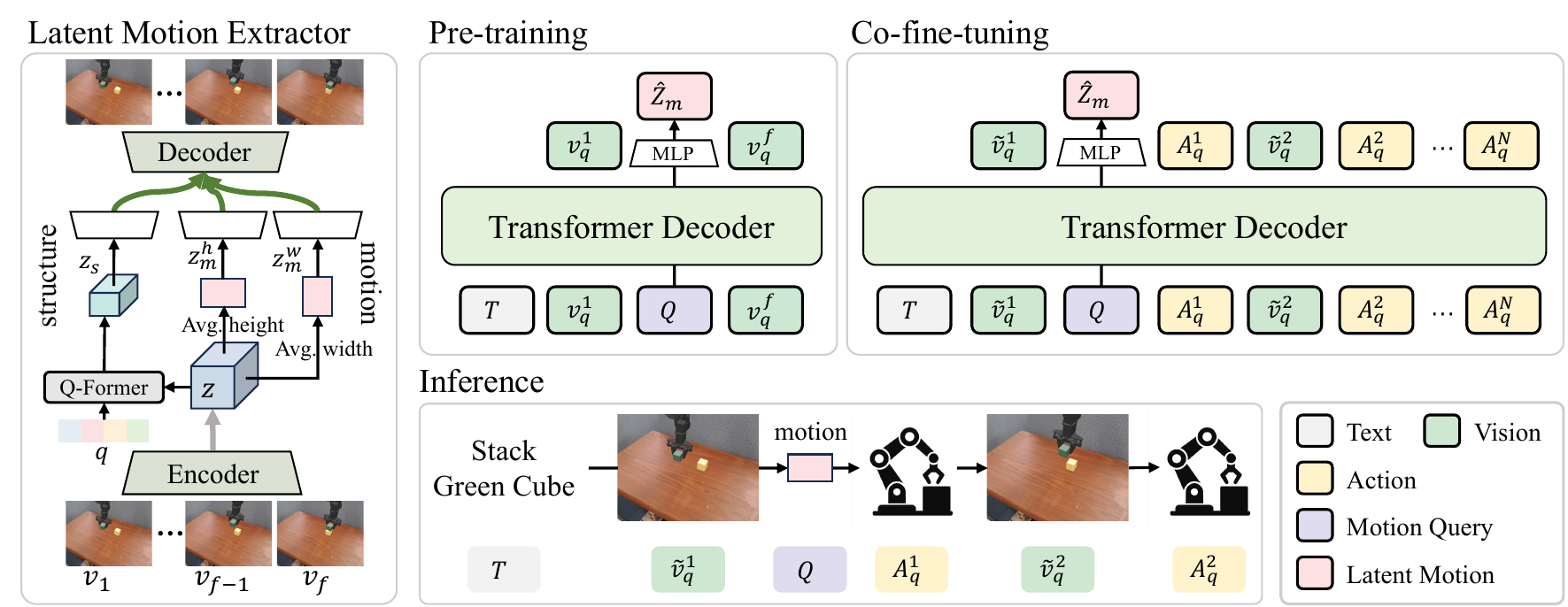}
 \caption{
\textbf{Overview of the CoWVLA framework.}
CoWVLA consists of two core components: a latent motion extractor and a VLA decoder.
The latent motion extractor, implemented as a video VAE, disentangles each video segment into a structure latent $z_s$ and two directional motion latents $z_m^h$ and $z_m^w$, which are concatenated into a unified latent motion vector $z_m$.
The VLA decoder performs unified autoregressive modeling over multimodal sequences.
During pre-training, the model takes the instruction and initial frame as input, and uses a learnable motion query $Q$ to predict the latent motion $\hat{z}_m$ while reconstructing the terminal frame of the video segment.
During co-fine-tuning, the input expands into alternating keyframe–action pairs; $Q$ continues to aggregate temporally continuous latent dynamics, guiding multi-step action generation under sparse visual observations.
}
   \label{fig:architecture}
   \vspace{-3ex}
\end{figure*}

\textbf{World Models for Robotics.}
World models are commonly employed to capture environment states and their future evolution, and have been widely applied in areas such as autonomous driving~\cite{wang2024driving,wei2024occllama}, image and video generation~\cite{bruce2024genie,lin2024opensora,wu2024ivideogpt,wang2024towards,sun2024eggen,di2025dh}, and robotics~\cite{wang2025unified,cen2025worldvla,assran2025vjepa,routray2026vipra,gao2025adaworld,zhang2025dreamvla}.
When combined with VLA models, most approaches~\cite{wu2024gr_1,cheang2024gr_2,zhao2025cot,cen2025worldvla,wang2025unified,zhong2025flowvla} rely on predicting future visual states to provide implicit world knowledge and demonstrate improved performance in robotic manipulation.
UVA~\cite{li2025unified} further jointly optimizes video prediction and action prediction using diffusion models, enhancing both visual reasoning and control inference efficiency.
However, these methods require reconstructing full visual frame sequences, leading to high computational cost and heavy resource consumption.

\textbf{Latent Actions for Robotics.}
Latent-action methods learn a compact latent transition between two frames to model environment dynamics.
LAPA~\cite{ye2025lap} introduces a three-stage framework (including latent action quantization, latent pretraining, and action fine-tuning), leveraging large-scale pseudo-action supervision to improve learning of real-world robotic control. MoTo~\cite{chen2025moto} follows this paradigm with enhancements in motion quantization and real action quality. 
TLA~\cite{bu2025tla} further disentangles task-relevant and task-irrelevant motion factors.
However, these approaches generally restrict latent action modeling to frame pairs, limiting their ability to capture long-range temporal dynamics. Although Villa-X~\cite{chen2025villa} extends latent actions to multi-frame settings, it still generates one latent action per local frame pair, resulting in limited temporal consistency. 
Moreover, the latent action representations inevitably encode static appearance and contextual details. 
While TLA~\cite{bu2025tla} mitigates this issue by decoupling task relevance, an ideal latent space should explicitly separate structure from motion, producing cleaner and more interpretable action representations.

\textbf{Video Compression and Decoupling.}
Recent methods in video representation learning have increasingly focused on compressing visual information into disentangled latent spaces that separately encode spatial structure and temporal motion~\cite{wu2024ivideogpt,shi2024motion,lew2025disentangled,yu2024efficient,wang2025vidtwin}.
The design of our latent motion space is inspired by these advances.
Models like CMD~\cite{yu2024efficient} and VidTwin~\cite{wang2025vidtwin} have successfully disentangled overall content and dynamic information in a highly compressed latent space.
This factorization provides a compact, continuous, and meaningful representation of how scenes evolve. 
While these models were developed for video generation, we are the first to hypothesize and demonstrate that their pretrained latent motion space can serve as a powerful dynamic prior for a robotic world model.

\section{Method}

\subsection{Overall Framework}

We consider a robotic manipulation task that involves executing a sequence of actions conditioned on a language instruction and visual observations.
The instruction is denoted as $T$. 
The raw action sequence is $\mathbf{A}_{1:t}=\{a_1,\ldots,a_t\}$. 
To enable discrete sequence modeling, the action sequence $\mathbf{A}_{1:t}$ is partitioned into consecutive chunks of fixed length $l_a$, i.e.,
$\mathbf{A}_{1:t} = \bigcup_{j=1}^{N} \mathbf{A}^{j}, \quad 
\mathbf{A}^{j} = \{a_{(j-1)l_a+1}, \ldots, a_{jl_a}\}$,
and each chunk $\mathbf{A}^{j}$ is then quantized into a discrete token sequence
$\mathbf{A}_q^{j}$,
using the FAST~\cite{pertsch2025fast} algorithm.
The raw corresponding visual observation sequence is represented as $\mathbf{V}_{1:t}=\{v_1,\ldots,v_t\}$, where each frame $v_i\in\mathbb{R}^{H\times W\times 3}$.
We extract the first frame of each action chunk as a keyframe:
$\tilde{\mathbf{V}}
= \{\tilde{v}_j\}_{j=1}^N
= \{v_{(j-1)l_a+1}\}_{j=1}^N,$
where each $\tilde{v}_j$ is subsequently quantized into a visual token $\tilde{v}_q^{j}$ using VQGAN~\cite{esser2021vqgan}.
Additionally, a learnable motion query token $Q\in\mathbb{R}^{D_Q}$ is introduced as a world dynamics query, whose hidden representation summarizes past context and provides a future dynamics-aware conditioning signal for generating subsequent vision or action tokens. 

The overall framework consists of two models and three training stages. 
The first model is the latent motion extractor (video VAE paradigm), which encodes a video sub-sequence $\mathbf{V}_{1:f}$ into an intermediate latent $z\in\mathbb{R}^{d_z\times f\times h\times w}$, and decomposes it into a structural feature $z_s$ and two directional motion features $z_{m}^h$ and $z_{m}^w$. 
The two motion components are concatenated to form a unified latent motion vector $z_m\in\mathbb{R}^{D_m}$, providing the ground-truth supervision. 
The second model is the VLA decoder (Transformer-decoder paradigm), which performs unified autoregressive next-token prediction across modalities. 
During pre-training, the input sequence is organized as $[T, v_q^1, Q, v_q^f]$.
The final hidden representation corresponding to the query token $Q$, obtained from the VLA decoder, is fed into an MLP to predict the latent motion $\hat{z}_m$.
This stage enables the model to infer latent dynamics and future observations from language and the initial visual input.
During co-fine-tuning, we use alternating keyframes and action tokens, e.g., $[T, \tilde{v}_q^{1}, Q, \mathbf{A}_q^1, \tilde{v}_q^{2}, \mathbf{A}_q^2, \ldots]$. 
The model continues to predict a latent motion vector $\hat{z}_m$ at $Q$ position. As a result, the model maintains explicit dynamics reasoning under sparse keyframe observations and generates stable multi-step actions from compact latent representations.

\subsection{Latent Motion Extractor}

To encode temporal dynamics in a compact latent space, we adopt a pretrained video variational autoencoder~\cite{wang2025vidtwin} as the latent motion extractor.
As illustrated in Figure~\ref{fig:architecture}, the extractor achieves structure--motion disentanglement through two dedicated branches.
Given a video segment $\mathbf{V}_{1:f}$, the encoder produces a latent tensor
$z \in \mathbb{R}^{d_z \times f \times h \times w}.$
The structure branch employs a Q-Former~\cite{li2023blip} module with a set of learnable queries $\{q_i\}_{i=1}^{n_q}$ to aggregate global semantics and low-frequency dynamics along the temporal dimension, yielding
$z_s \in \mathbb{R}^{d_s \times n_q \times h_s \times w_s}, n_q \leq f.$
The motion branch operates along spatial dimensions: several convolutional layers reduce the dimension of $z$ and produce $z' \in \mathbb{R}^{d_m \times f \times h_m \times w_m}$.
Then, spatial averaging $\mu(\cdot)$ is applied independently along the height and width axes to extract directional motion embeddings:
$z_m^{h} = \mu_{h}(z') \in \mathbb{R}^{d_m \times f \times w_m}, 
z_m^{w} = \mu_{w}(z') \in \mathbb{R}^{d_m \times f \times h_m}.$
These two motion components are concatenated and flattened to form a unified latent motion representation:
$z_m \in \mathbb{R}^{D_m}, D_m=f \times d_m \times (h_m+w_m).$
In the decoder stage, the three latent components $(z_s, z_m^h, z_m^w)$ are upsampled through convolutional and MLP layers to the same spatial and temporal size, summed together, and then fed into the decoder to reconstruct $\hat{\mathbf{V}}_{1:f}$.
The training objective follows the original VAE design~\cite{wang2025vidtwin}, combining reconstruction loss $\mathcal{L}_{\text{rec}}$, perceptual loss $\mathcal{L}_{p}$, adversarial loss $\mathcal{L}_{\text{GAN}}$, and KL-divergence regularization loss $\mathcal{L}_{\text{KL}}$ to preserve temporal consistency and visual realism:
\begin{equation}
\mathcal{L}_{vae}
= \mathcal{L}_{\text{rec}}
+ \lambda_{p}\mathcal{L}_{p}
+ \lambda_{\text{GAN}}\mathcal{L}_{\text{GAN}}
+ \lambda_{\text{KL}}\mathcal{L}_{\text{KL}}.
\end{equation}

Through explicit structure–motion disentanglement and mild adaptation, the extractor yields a compact, interpretable, and transferable latent representation well-suited for robotic scenarios, providing effective supervision for downstream VLA pre-training and co-fine-tuning.

\subsection{Pre-training to Think in Latent Motion}

The pre-training stage aims to align language and initial visual observations with latent motion representations, enabling the model to reason about continuous temporal dynamics in the latent space and predict the terminal frame of the video segment.
Given a continuous video segment $\mathbf{V}_{1:f} = \{v_1, \ldots, v_f\}$, the latent motion extractor produces a latent motion supervision signal $z_m$.
Its first and last frames are quantized into discrete visual tokens, denoted as $v_q^1$ and $v_q^f$, respectively.
Based on this, we organize the input sequence to the VLA decoder as: $[T, v_q^1,  Q,  v_q^f],$
where $T$ denotes the instruction, $v_q^1$ represents the initial observation, $Q$ is a learnable motion query token, and $v_q^f$ corresponds to the visual state that would be reached after applying the underlying motion from $v_1$ through $z_m$.
During the forward pass, the hidden state at the query position is fed to an MLP to predict the latent motion $\hat{z}_m$.

To prevent information leakage, causal masking is applied so that $Q$ only attends to $\{T, v_q^1\}$ while being masked from $v_q^f$.
The training objective contains latent motion supervision and terminal-frame visual consistency:
\begin{equation}
\mathcal{L}_{\text{pretrain}}
=
 \|\hat{z}_m - z_m\|_2^2
+
\sum_{x \in \{1,f\}} \mathrm{CE}(\hat{v}_q^x, v_q^x),
\end{equation}
where the first term enforces that the latent representation extracted at $Q$ accurately summarizes the continuous motion from $v_1$ to $v_f$, while the second ensures that the model forms a coherent prediction of the resulting future state.
Through this stage, the model learns to infer latent temporal dynamics directly from language and the initial frame, thus establishing a dynamics-aware prior for subsequent action modeling.

\subsection{Co-Fine-Tuning for Aligning Latent Dynamics with Action Policies}

After the pre-training stage establishes a dynamics-aware prior in the latent motion space, the co-fine-tuning stage further aligns latent motion reasoning with discrete action modeling in a unified autoregressive framework, enabling stable multi-step control under sparse keyframe observations.
Given a continuous video sequence $\mathbf{V}_{1:f}$ and its corresponding action sequence $\mathbf{A}_{1:f}$, we extract $N=f/l_a$ keyframes and quantize them into visual tokens:
$
\tilde{\mathbf{V}}_q=\{\tilde{v}_q^{1},\ldots,\tilde{v}_q^{N}\},
$
where $\tilde{v}_q^j = v_q^{(j-1)l_a+1}$.
We further quantize the action sequence using FAST~\cite{pertsch2025fast}:
$
\mathbf{A}_{1:f}\ \xrightarrow{\text{FAST}}\ \{\mathbf{A}_q^{1},\ldots,\mathbf{A}_q^{N}\}.
$
The input sequence adopts a ``single-$Q$ for the full window'' design:
$
[T,\ \tilde{v}_q^{1},\ Q,\ \mathbf{A}_q^{1},\ \tilde{v}_q^{2},\ \mathbf{A}_q^{2},\ \ldots,\ \mathbf{A}_q^{N}],
$
where the query token $Q$ appears only once after the first keyframe and serves as a latent dynamics aggregator for the entire temporal horizon. The decoder autoregressively predicts both action and visual tokens; the hidden state at $Q$ is passed through an MLP to produce a single latent motion vector $\hat{z}_m$, enforcing consistency between latent dynamics and subsequent predictions. 
As in pre-training, causal masking prevents $Q$ from attending to future keyframes and actions, compelling the model to reason over latent dynamics rather than directly peeking at future states.

The co-fine-tuning objective consists of three terms:
\begin{equation}
\begin{aligned}
\mathcal{L}_{\text{finetune}}
= &
\sum_{j=1}^{N}
\mathrm{CE}\!\left(\hat{\mathbf{A}}_q^{j},\ \mathbf{A}_q^{j}\right)
+
\lambda_1
\left\|\hat{z}_m - z_m(\mathbf{V}_{1:f})\right\|_2^2
\\
&+
\lambda_2
\sum_{j=1}^{N}
\mathrm{CE}\!\left(\hat{\tilde{v}}_q^{j},\ \tilde{v}_q^{j}\right).
\end{aligned}
\end{equation}
Here, $z_m(\mathbf{V}_{1:f})$ is a continuous latent motion supervision signal produced by the pretrained extractor. 
The first term ensures accurate execution of discrete actions. 
The second term encourages the latent representation at the query token to faithfully capture the continuous dynamics from $v_1$ to $v_f$. 
The third term anchors motion predictions to sparse visual checkpoints, maintaining consistent state transitions driven by the predicted dynamics.
\begin{table*}[t]
    \caption{Comparison of different methods on the LIBERO~\cite{liu2023libero} and SimplerEnv-WidowX~\cite{li2024evaluating} benchmarks.
    The best and the second-best values for each metric are bold and \underline{underlined} respectively. }
    \label{tab:eval_combined}
    \centering
    \scalebox{0.87}{
    \begin{tabular}{lccccc|cccccc}
    \toprule
    \multirow{2}{*}{\textbf{Model}} & \multicolumn{5}{c|}{\textbf{LIBERO}} & \multicolumn{5}{c}{\textbf{SimplerEnv-WidowX}} \\
    \cmidrule(lr){2-6} \cmidrule(lr){7-11}
     & SPATIAL & OBJECT & GOAL & LONG & \textbf{Avg.} 
    & Stack Block & Put Carrot & Put Spoon & Put Eggplant & \textbf{Avg.} \\
    \midrule
    OpenVLA~\cite{kim2024openvla} & 0.849 & 0.884 & 0.792 & 0.537 & 0.765 & 0.000 & 0.000 & 0.000 & 0.041 & 0.010 \\
    SpatialVLA~\cite{qu2025spatialvla} & 0.882 & 0.899 & 0.786 & 0.555 & 0.781 & 0.292 & 0.250 & 0.167 & \textbf{1.000} & 0.427 \\
    CogACT~\cite{li2024cogact} & 0.960 & 0.874 & 0.868 & 0.846 & 0.887 & 0.150 & 0.508 & 0.717 & 0.675 & 0.513 \\
    Dita~\cite{hou2025dita} & 0.842 & 0.963 & 0.854 & 0.638 & 0.824 & – & – & – & – & – \\
    $\pi_0$~\cite{black2024pi_0} & 0.968 & 0.988 & 0.958 & 0.852 & 0.942 & 0.167 & 0.000 & 0.291 & 0.625 & 0.401 \\
    $\pi_0$-FAST~\cite{pertsch2025fast} & 0.964 & 0.968 & 0.886 & 0.602 & 0.855 & 0.108 & 0.219 & 0.291 & 0.666 & 0.483 \\
    GR00T N1~\cite{bjorck2025gr00t} & 0.944 & 0.976 & 0.930 & 0.906 & 0.939 & 0.167 & 0.458 & 0.625 & 0.208 & 0.495 \\
    \midrule
    \textbf{w/ Latent Actions}\\
    LAPA~\cite{ye2025lap} & – & – & – & – & – & 0.542 & 0.458 & 0.708 & 0.583 & 0.573 \\
    villa-X~\cite{chen2025villa} & \textbf{0.975} & 0.970 & 0.915 & 0.745 & 0.901 & \underline{0.613} & 0.463 & 0.779 & 0.646 & 0.625 \\
    TLA~\cite{bu2025tla} & 0.965 & 0.968 & \textbf{0.956} & \underline{0.920} & \underline{0.952} & 0.028 & 0.556 & 0.528 & 0.806 & 0.480 \\
    \midrule
    \textbf{w/ World Model}\\
    WorldVLA~\cite{cen2025worldvla} & 0.856 & 0.890 & 0.826 & 0.590 & 0.791 & – & – & – & – & – \\
    CoT-VLA~\cite{zhao2025cot} & 0.875 & 0.916 & 0.876 & 0.690 & 0.811 & – & – & – & – & – \\
    UniVLA~\cite{wang2025unified} & 0.960 & \textbf{0.992} & 0.932 & 0.914 & 0.950 & 0.292 & \underline{0.625} & \textbf{0.833} & \textbf{1.000} & 0.687 \\
    FlowVLA~\cite{zhong2025flowvla} & 0.932 & 0.950 & 0.916 & 0.726 & 0.881 & \textbf{0.625} & \underline{0.625} & 0.708 & \textbf{1.000} & \underline{0.740} \\
    \midrule
    \rowcolor{gray!20}
    Ours & \underline{0.972} & \underline{0.978} & \underline{0.946} & \textbf{0.928} & \textbf{0.956} 
    & \textbf{0.625} & \textbf{0.667} & \underline{0.792} & \underline{0.958} & \textbf{0.760} \\
    \bottomrule
    \end{tabular}}
\end{table*}

\section{Experiments}
\subsection{Benchmarks}

\textbf{LIBERO.}
The LIBERO~\cite{liu2023libero} benchmark is designed for studying knowledge transfer in multitask and lifelong robot learning, requiring both \textit{declarative knowledge} about objects and spatial relations and \textit{procedural knowledge} about motion and behaviors. 
It contains four task suites: LIBERO-Spatial emphasizes spatial reasoning by placing a bowl based on its location, LIBERO-Object focuses on object recognition via picking and placing distinct objects, LIBERO-Goal tests procedural learning with varying task goals under fixed objects, and LIBERO-Long contains ten long-horizon tasks with diverse objects, layouts, and goals.

\textbf{SimplerEnv.}
SimplerEnv~\cite{li2024evaluating} is a collection of manipulation evaluation environments for common real-world robot setups, showing strong correlation with real-robot performance. It enables assessing the transferability and generalization of models trained on real-world video data. We evaluate on four tasks using a 7-DoF WidowX robotic arm.

\subsection{Implementation Details}

Our latent motion extractor is built upon a pretrained video VAE (VidTwin~\cite{wang2025vidtwin}) and is further fine-tuned on a robot-centric dataset consisting of 237k videos (details provided in the appendix). 
Each video segment is uniformly sampled to 16 frames and resized to $224 \times 224$.
The structure latent $z_s$ has a shape of $4 \times 16 \times 7 \times 7$, 
while the directional motion embeddings $z_m^{h}$ and $z_m^{w}$ have shapes of $8 \times 16 \times 7$.
The motion latent dimension is $D_m = 1792$.

The backbone of our VLA model follows the design of UniVLA~\cite{wang2025unified} and is based on the 8.5B-parameter VLM Emu3~\cite{wang2024emu3}. 
Visual observations are quantized into discrete tokens using VQGAN~\cite{esser2021vqgan}, while actions are partitioned into chunks and discretized into tokens using the FAST algorithm~\cite{pertsch2025fast}.
During the pre-training stage, we trained the model using the aforementioned 237k videos with pretrained Emu3 initialization. 
From each video, we extracted a frame sequence of length $f=16$, where the first and last frame tokens supervise visual modeling, and the latent motion extracted from VidTwin provides supervision. 
We trained using a batch size of 256 for 10k steps.
During the co-fine-tuning stage, we initialized from the pretrained checkpoint and trained on the benchmark-specific datasets.
For the LIBERO benchmark, we used the mixed data from the four task suites curated by OpenVLA~\cite{kim2024openvla}, including both third-person and wrist-mounted views. 
We trained the model with a batch size of 128 for 8k iterations, resized all images to $200 \times 200$, set the action chunk length to $l_a = 10$, and used $\lambda_1 = 0.1$ and $\lambda_2 = 0.01$.
For SimplerEnv, we trained the model on the Bridge V2 dataset~\cite{walke2023bridgev2} with a batch size of 128 for 12k iterations. 
Single-view images were resized to $256 \times 256$, the action chunk length is set to $l_a = 5$, and we used $\lambda_1 = 0.1$ and $\lambda_2 = 0$.
In the co-fine-tuning stage, we set $N=2$, where two visual observations and two corresponding ground-truth action chunks were used.

Further training details and supplementary results are provided in the appendix.

\begin{table*}[t]
    \caption{Evaluation of VAE-Reconstructed Videos and downstream fine-tuning performance on SimplerEnv-WidowX~\cite{li2024evaluating}.}
    \label{tab:vae_recon_video}
    \centering
    \scalebox{0.96}{
    \begin{tabular}{lccc|ccccc}
    \toprule
    \multirow{2}{*}{Model} & \multicolumn{3}{c|}{Reconstruction Metrics} & \multicolumn{5}{c}{Simulation Evaluation} \\ 
    \cmidrule(lr){2-4} \cmidrule(lr){5-9}
     & PSNR$\uparrow$ & SSIM$\uparrow$ & LPIPS$\downarrow$ & Stack Block & Put Carrot & Put Spoon & Put Eggplant & \textbf{Average} \\
    \midrule
    Pretrain & 32.7 & 0.923 & \textbf{0.122} & 0.458 & \textbf{0.750} & 0.792 & 0.917 & 0.729 \\
    Finetune & \textbf{33.4} & \textbf{0.934} & 0.123 & \textbf{0.625} & 0.667 & \textbf{0.792} &\textbf{ 0.958} & \textbf{0.760} \\
    \bottomrule
    \end{tabular}
    }
\vspace{-2ex}
\end{table*}

\begin{figure}[t]
  \centering   
\includegraphics[width=1\linewidth]{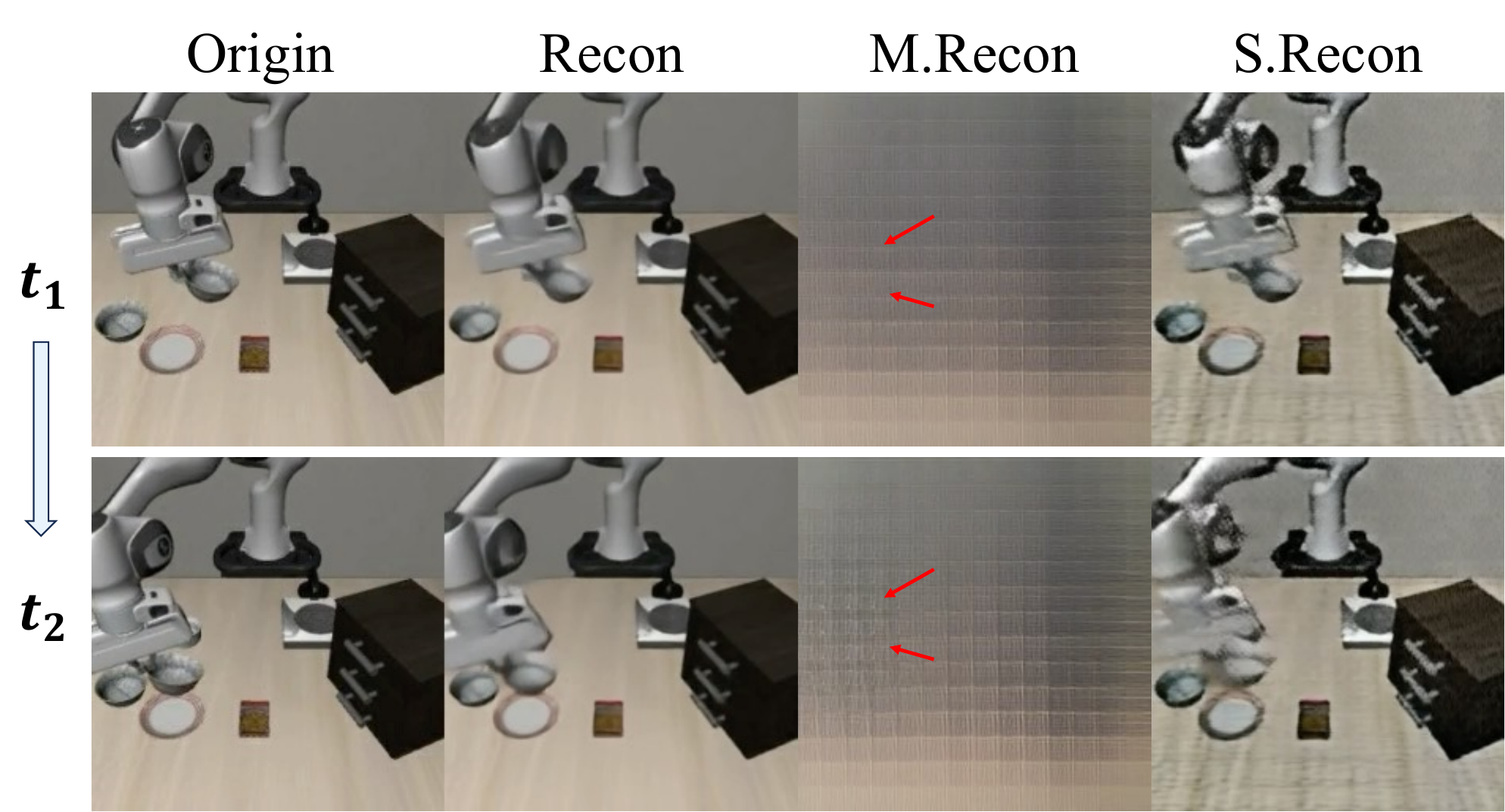}
   \caption{
Visualization of the disentangled motion and structure latents.
We select two frames ($t_1$ and $t_2$) and show the original (Orig.) and reconstructed (Recon.) frames.
``M. Recon.'' and ``S. Recon.'' denote the reconstructions obtained by decoding only the motion latent or only the structure latent, respectively.
The structure latent preserves the global scene layout, whereas the motion latent captures motion and fine-grained temporal details.
}
\label{fig:motion_vis_S_D_latent}
\vspace{-14pt}
\end{figure}

\subsection{Comparison with SOTA Methods}
We compared CoWVLA against three representative categories of methods: VLA baselines (OpenVLA~\cite{kim2024openvla}, SpatialVLA~\cite{qu2025spatialvla}, CogACT~\cite{li2024cogact}, DiTA~\cite{hou2025dita}, $\pi_0$~\cite{black2024pi_0}, $\pi_0$-FAST~\cite{pertsch2025fast}, GR00T-N1~\cite{bjorck2025gr00t}), latent-action approaches (LAPA~\cite{ye2025lap}, villa-X~\cite{chen2025villa}, TLA~\cite{bu2025tla}), and world-model approaches (WorldVLA~\cite{cen2025worldvla}, CoT-VLA~\cite{zhao2025cot}, UniVLA~\cite{wang2025unified}, FlowVLA~\cite{zhong2025flowvla}). 
These methods respectively model: (i) actions directly, (ii) frame-to-frame latent transitions, and (iii) pixel/token-level future frames.
They collectively represent the main paradigms in current VLA pretraining and provide strong and fair comparison points. 
The results are shown in Table~\ref{tab:eval_combined}.

Overall, our CoWVLA achieves SOTA performance with superior cross-domain robustness. 
We observe that TLA achieves a strong 0.952 on LIBERO but significantly drops to 0.480 on SimplerEnv, while FlowVLA is strong on SimplerEnv (0.740) but noticeably weaker on LIBERO (0.881). UniVLA shows a more balanced performance (0.950/0.698). 
In contrast, CoWVLA achieves 0.956/0.760 on the two benchmarks, outperforming UniVLA on both and demonstrating higher absolute performance and greater cross-domain stability.

\subsection{Latent Motion Analysis}

In this subsection, we analyze the effectiveness of the proposed disentangled latent space from three perspectives: the separation of structure and motion factors, the improved adaptiveness of the motion latent after fine-tuning on robot data, and the enhanced capability of modeling future dynamics. These results collectively verify that our latent motion representation provides a clearer physical prior and stronger action reasoning ability.

\begin{figure}[t]
  \centering   
\includegraphics[width=\linewidth]{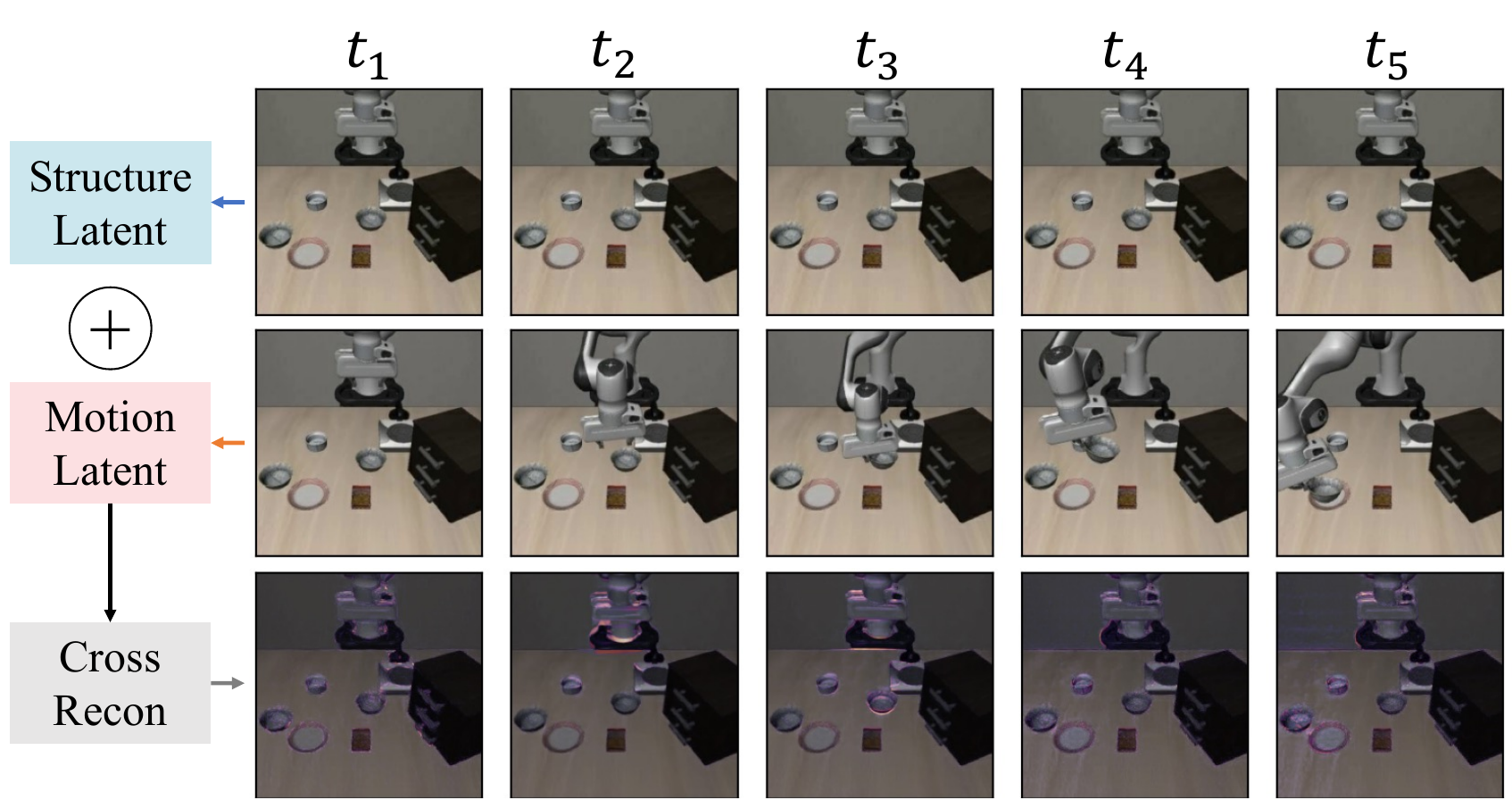}
   \caption{
Cross-reconstruction visualization. 
We extract the structure latent from the static video in the first row and the motion latent from the robot-arm motion video in the second row. 
By combining the two latents, we reconstruct the video shown in the third row. 
We compute the difference between the cross-reconstructed frames and the static frames to highlight the changed regions, which correspond to the robot arm's motion.
}
\label{fig:motion_structure_heatmap}
\vspace{-10pt}
\end{figure}

\textbf{Effective decoupling of structure and motion latent.}
As shown in Figure~\ref{fig:motion_vis_S_D_latent}, we reconstruct frames using only the motion latent (M. Recon.) or only the structure latent (S. Recon.). 
The structure latent preserves global scene layout and object appearance, whereas the motion latent captures robot arm trajectories and fine-grained temporal dynamics. 
Figure~\ref{fig:motion_structure_heatmap} provides additional evidence through cross-reconstruction.
Since motion cues are subtle in individual frames, we visualize the pixel-wise differences, which highlight the motion-affected regions and show that injecting the motion latent alters only the dynamic parts while keeping the static structure intact.
These visualizations demonstrate that our latent space effectively separates content structure and dynamic information, providing a more interpretable representation for downstream visuomotor reasoning.

\textbf{Fine-tuning on robot data improves motion latent quality.}
As presented in Table~\ref{tab:vae_recon_video}, fine-tuning the latent motion extractor on robot data not only improves reconstruction quality (higher PSNR and SSIM) but also boosts downstream performance. In the SimplerEnv-WidowX evaluation, the average task success rate increases from 0.729 to 0.760. This confirms that motion latents adapted to the robot domain contain higher-quality dynamic cues that benefit policy learning.

\begin{figure}[t]
  \centering   
\includegraphics[width=\linewidth]{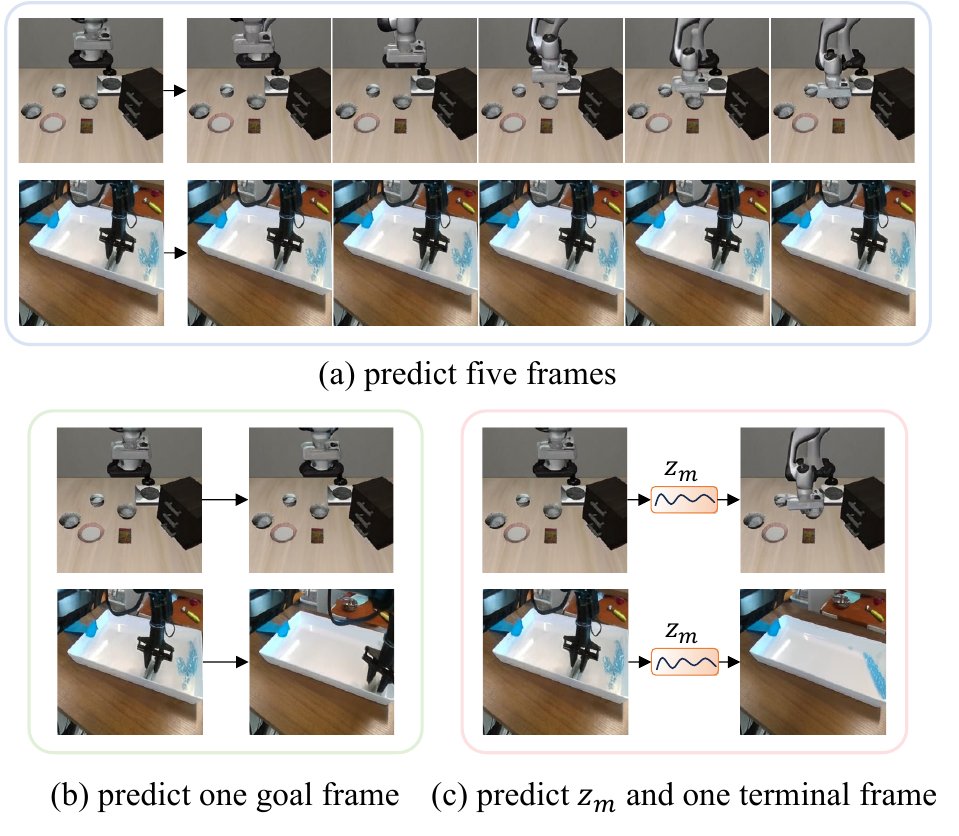}
   \caption{
Comparative visualization of future-frame prediction strategies.
There are two tasks demonstrated: i) pick up the black bowl from the table center and place it on the plate, and ii) sweep into a pile.
(a) The world-model approach predicts five future frames.
(b) The single-goal-frame approach predicts one goal frame.
(c) Our method reasons through a learned motion latent $z_m$, producing more reasonable and instruction-aligned frames.
   }
\label{fig:univla_cotvla_ours}
\end{figure}

\textbf{Motion latent enhances dynamic modeling for future frame prediction.}
As illustrated in Figure~\ref {fig:univla_cotvla_ours}, we visualize the future frame predictions under different pretraining strategies. 
From top to bottom, the tasks in each subfigure are:
i) pick up the black bowl from the table center and place it on the plate,
ii) sweep into a pile.
World-model-based approaches reconstruct redundant background pixels and therefore struggle to focus on interactive motion, while single-goal-frame prediction lacks supervision of temporal evolution and often produces unstable goal frames. 
This leads both strategies to easily generate results with no changes, such as Figure~\ref{fig:univla_cotvla_ours} (b) Task i.
In contrast, our model leverages the motion latent as a ``chain of world'' during reasoning, achieving physically plausible future states that align more closely with the instructions.

\subsection{Ablation and Efficiency Analysis}

\begin{figure}[t]
  \centering   
\includegraphics[width=0.85\linewidth]
{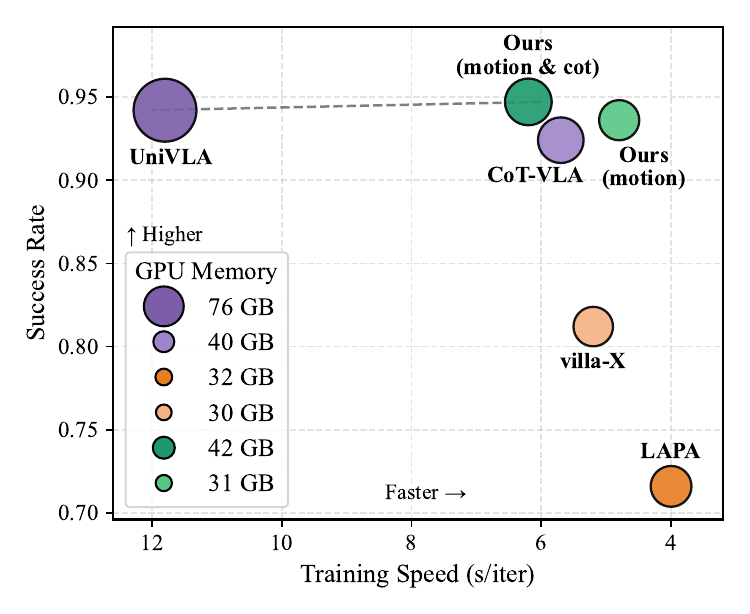}
\vspace{-1ex}
   \caption{
   Comparison of pre-training efficiency and task performance on LIBERO~\cite{liu2023libero} across different methods.
   Blue and orange circles denote world-model and latent-action baselines, respectively, while green circles denote our configurations. 
   Circle size indicates training-time GPU memory usage.
   Our method balances pre-training efficiency and performance, achieving a higher success rate with moderate computational efficiency.
}
\label{fig:efficiency_tradeoff_groups}
\vspace{-2ex}
\end{figure}

In this section, we conduct an in-depth analysis of key modules, hyperparameter settings, and training efficiency. 
Experiments in Table~\ref{tab:ablation_libero} and Table~\ref{tab:ablation_libero_loss_weight} adhere to a unified dataset and training configuration, with a batch size of 256 for 10k steps during the pre-training phase and a batch size of 128 for 8k steps during the co-fine-tuning phase. 
In Table~\ref{tab:ablation_libero}, we provide a unified comparison of the effectiveness of latent action, world model, and our proposed method. 
In Table~\ref{tab:ablation_libero_loss_weight}, we analyze the effect of the loss weighting ratio between the latent motion loss ($\lambda_1$) and the visual token loss ($\lambda_2$) on task success rates during the co-fine-tuning strategy.
In addition, we analyze the pre-training cost and task success rate of different methods in Figure~\ref{fig:efficiency_tradeoff_groups}.
The main conclusions are as follows.

\begin{table}[t]
\centering
\caption{Ablation study on the LIBERO~\cite{liu2023libero} benchmark.}
\label{tab:ablation_libero}
\scalebox{0.71}{
\begin{tabular}{llccccc}
\toprule
Config & Variant & Spatial & Object & Goal & Long & \textbf{Average} \\
\midrule
\multirow{5}{*}{Latent Action} 
 & w/o LA & 0.622 & 0.146 & 0.694 & 0.328 & 0.448\\
 & LAPA style & 0.718 & 0.852 & 0.804 & 0.488 & 0.716 \\
 & villa-X style & 0.840 & 0.904 & 0.834 & 0.668 & 0.812  \\
 & structure latent & 0.856 & 0.898 & 0.822 & 0.692 & 0.817 \\
 & motion latent & 0.916 & 0.932 & 0.886 & 0.774 & 0.877  \\
\midrule
\multirow{2}{*}{World Model} 
 & UniVLA Style & 0.958 & 0.978 & 0.932 & 0.898 & 0.942 \\
 & CoT-VLA style & 0.942 & 0.964 & 0.950 & 0.838 & 0.924 \\
 \midrule
 \multirow{2}{*}{Ours}  &  motion & \textbf{0.960} & \textbf{0.980} & 0.922 & 0.882 & 0.936 \\
& motion \& cot  & 0.948 & 0.974 & \textbf{0.958} & \textbf{0.906} & \textbf{0.947} \\
\bottomrule
\end{tabular}
}
\vspace{-2ex}
\end{table}

\romannumeral 1) Our latent motion modeling significantly outperforms existing latent action methods.
The ``Latent Action'' part of Table~\ref{tab:ablation_libero} compares several baselines. 
The ``w/o LA'' variant, which skips pre-training and fine-tunes directly on LIBERO data, achieves the lowest average success rate (0.448). ``LAPA style'' (0.716) and ``villa-X style'' (0.812) both outperform the ``w/o LA'' variant, with ``villa-X style'' achieving stronger performance by modeling richer multi-frame information. 
Our method separates the latent into a ``structure latent'' (0.817) capturing content and texture, and a ``motion latent'' (0.877) encoding dynamic information. 
Modeling with the cleaner motion notably improves task success rate.

\romannumeral 2) World model methods show stronger overall performance than latent action methods.
In the ``World Model'' part of Table~\ref{tab:ablation_libero}, both ``UniVLA style'' (pretrained with six frames) and ``CoT-VLA style'' (pretrained with initial and target frames) achieve higher success rates (0.942 and 0.924, respectively) than those methods in the ``Latent Action'' category. 
Notably, ``UniVLA style'', which uses more frames, performs better, indicating that world model methods have a distinct advantage in temporal modeling and learning knowledge of environmental evolution.

\romannumeral 3) Our method achieves superior performance to latent action and world models.
The ``Ours'' part in Table~\ref{tab:ablation_libero} presents two configurations of our method. Both use latent motion supervision during pre-training and set $\lambda_1=0.1, \lambda_2=0$ during fine-tuning (i.e., using only real action and latent motion losses). The ``motion'' configuration does not use the final frame $v_f$ during pre-training and achieves a success rate of 0.936. 
In contrast, the ``motion \& cot'' configuration adds supervision from $v_f$ during pre-training and improves the success rate to 0.947. 
This yields two conclusions: first, introducing latent motion during the fine-tuning phase effectively guides the inference of real actions; second, introducing $v_f$ as an evolutionary target during pre-training significantly enhances the model's perception and understanding of environmental evolution.

\begin{table}[t]
\centering
\caption{Ablation study of loss weights on the LIBERO~\cite{liu2023libero} benchmark.}
\label{tab:ablation_libero_loss_weight}
\scalebox{0.94}{
\begin{tabular}{cc|ccccc}
\toprule
$\lambda_1$ & $\lambda_2$ & Spatial & Object & Goal & Long & \textbf{Average} \\
\midrule
0.0 & 0.0 & 0.922 & 0.962 & 0.862 & 0.742 & 0.872 \\
0.1 & 0.0 & 0.960 & 0.980 & 0.922 & 0.882 & 0.936 \\
1.0 & 0.0 & 0.958  & 0.970 & 0.950 & 0.902 & 0.945 \\
0.1 & 0.05 & 0.954 & 0.972 & 0.944 & 0.914 & 0.946 \\
0.1 & 0.01 & 0.970 & 0.964 & 0.958 & 0.926 & \textbf{0.955} \\
1.0 & 0.01 & 0.970 & 0.956 & 0.934 & 0.922 & 0.946 \\
\bottomrule
\end{tabular}
}
\vspace{-2ex}
\end{table}

\romannumeral 4) Balancing latent motion and visual token losses during co-fine-tuning further improves performance.
Table~\ref{tab:ablation_libero_loss_weight} presents an ablation study on the loss weights $\lambda_1$ (latent motion) and $\lambda_2$ (visual token) during the co-fine-tuning stage, based on the same pretrained model.
First, we fix $\lambda_2=0$ to analyze the impact of $\lambda_1$. When $\lambda_1=0$ (no latent motion loss), the success rate is only 0.872. As $\lambda_1$ increases from 0.1 to 1.0, the success rate improves from 0.936 to 0.945, indicating that the guiding effect of latent motion is strengthening.
Next, we introduce the visual token loss $\lambda_2$.
By comparing ($\lambda_1=0.1, \lambda_2=0.05$) at 0.946 and ($\lambda_1=0.1, \lambda_2=0.01$) at 0.955, we find that the weight for visual token prediction should not be too high.
Then we tune $\lambda_1=1.0$ and $\lambda_2=0.01$, achieving an average success rate of 0.946.
This proves that simultaneously introducing latent motion ($\lambda_1=0.1$) and a low-weighted visual token prediction ($\lambda_2=0.01$) during the fine-tuning phase most effectively guides the inference of real actions.

\romannumeral 5) Our method balances pre-training efficiency and performance. 
As shown in Figure~\ref{fig:efficiency_tradeoff_groups}, we compare several methods from Table~\ref{tab:ablation_libero} in terms of training speed, GPU memory usage, and task success rate (batch size = 4 per GPU). 
UniVLA is the slowest and most memory-intensive, while LAPA is the fastest but less successful.
Our method has two configurations: ``motion'' without $v_f$ achieves the second-fastest speed and slightly lower performance than UniVLA, and ``motion \& cot'' with $v_f$ achieves a better balance of efficiency and performance, surpassing UniVLA in both.

\section{Conclusion}
In this work, we presented CoWVLA, which for the first time integrates the temporal reasoning capability of world models with a disentangled latent motion representation, enabling world modeling directly in a structure–motion separated latent space. 
By introducing the Chain-of-World paradigm, our method predicts a continuous latent motion chain and a terminal keyframe from the instruction and initial observation, compactly capturing temporal evolution and physical dynamics without reconstructing intermediate pixels. 
Extensive experiments on LIBERO and SimplerEnv benchmarks demonstrate that CoWVLA outperforms both world-model and latent-action approaches, while offering improved dynamic consistency and visuomotor grounding, thereby providing a more efficient pretraining route toward general-purpose robotic manipulation.

\textbf{Limitations}.
Despite its promising results, our approach still has limitations. 
The latent motion space remains dependent on the quality and domain coverage of the pretrained video VAE, which may introduce distribution mismatch in new environments. 
Moreover, the model relies on a large VLA backbone and substantial computational resources. 
We believe exploring more lightweight and scalable architectures, as well as further enhancing the coupling between latent dynamics and action learning, will broaden the applicability of our method to real-world robotics.

\section{Acknowledgments}
This work was supported by the National Natural Science Foundation of China (Grant No. 62277011), Project of Chongqing MEITC (Grant No. YJX-2025001001009), and CAAI-CANN Open Fund, developed on OpenI Community.

{
    \small
    \bibliographystyle{ieeenat_fullname}
    \bibliography{main}
}

\clearpage
\setcounter{page}{1}
\maketitlesupplementary
\setcounter{section}{0}
\setcounter{figure}{0}
\setcounter{table}{0}

\section{Implementation Details}
\subsection{Datasets}

We collected high-quality robot manipulation data for fine-tuning the Latent Motion Extractor (LME) and training the VLA, with the datasets summarized in Table~\ref{tab:dataset_counts}. Most of the data comes from the OXE~\cite{o2024openx} dataset, and we additionally include the Calvin~\cite{mees2022calvin} and Libero~\cite{liu2023libero} simulation datasets. For LME fine-tuning, we use only episode frames. In the VLA pre-training stage, we use both episode frames and text instructions. Following UniVLA~\cite{wang2025unified}, we adopt different sampling intervals for each dataset to ensure that the temporal gap between keyframes is approximately one second. We then uniformly sample 16 frames from the continuous frames covered by six keyframes for pre-training. Throughout this stage, only third-person view data is used, excluding wrist-camera views.

During the VLA co-fine-tuning stage, we train on the benchmark-specific training sets using text instructions, frames, and actions. For example, the BridgeV2 dataset~\cite{walke2023bridgev2} is used for the SimplerEnv-Bridge evaluation~\cite{li2024evaluating}, while the Libero~\cite{liu2023libero} evaluation uses the mixed data of four Libero task suites processed by OpenVLA~\cite{kim2024openvla}. In addition, the appendix includes extended experiments using the Fractal dataset~\cite{brohan2022rt1} for the Simpler-Google Robot~\cite{li2024evaluating} evaluation and the Calvin dataset~\cite{mees2022calvin} for the Calvin evaluation, covering both ABCD$\rightarrow$D and ABC$\rightarrow$D task settings. Across the co-fine-tuning experiments, Bridge and Google Robot training use only third-person views, while Libero and Calvin use both third-person and wrist views.

\subsection{Training Details}

For LME fine-tuning, we start from the VidTwin~\cite{wang2025vidtwin} pretrained model and fine-tune it on the video data from the datasets listed in Table~\ref{tab:dataset_counts}. 
We use 4 A800 GPUs with a per-GPU batch size of 4, randomly sampling 16 frames per video. 
Each frame is resized to 224$\times$224. 
The KL loss weight is set to 1e-6, and the reconstruction loss is reduced using the mean over all elements rather than the default reduction over the batch dimension only. 
We randomly sample 1000 videos from the training set as a validation set and select the checkpoint with the lowest reconstruction loss. 
The final model corresponds to the checkpoint trained for one epoch plus 20k iterations.

For VLA pre-training, we initialize from the 8.5B Emu3~\cite{wang2024emu3} pretrained checkpoint and train on the datasets in Table~\ref{tab:dataset_counts}.
The training is performed on 32 A800 GPUs with a per-GPU batch size of 8. 
Image observations are resized to 256$\times$256. 
We use the first and last frames of each video clip together with one learnable motion query, and the maximum sequence length is set to 2500 tokens. 
We train for 10k iterations in total, which takes roughly 24 hours.

For VLA co-fine-tuning, we follow the evaluation protocols from UniVLA~\cite{wang2025unified} for each benchmark. 
We load the checkpoint from the VLA pre-training stage and train with 16 A800 GPUs, using a batch size of 8 per GPU and full-parameter fine-tuning. 
The maximum sequence length is set to 3200 tokens. 
For SimplerEnv-Windowx~\cite{li2024evaluating}, we use BridgeV2~\cite{walke2023bridgev2} data with images resized to 256$\times$256 and train for 12k iterations. 
For SimplerEnv-Google Robot~\cite{li2024evaluating}, Fractal~\cite{brohan2022rt1} images are resized to 240$\times$192, and training continues for 16k iterations. 
For Libero~\cite{liu2023libero}, images are resized to 200$\times$200, and training runs for 8k iterations. 
For Calvin~\cite{mees2022calvin}, third-person views are resized to 200$\times$200 and wrist views to 80$\times$80, with training conducted for 12k iterations. 
The per-iteration training time across these configurations is similar; for example, Libero training takes about 25 hours for 8k iterations. 
Overall, each configuration requires roughly one to two days of training.

\begin{table}[t]
\caption{Training datasets.}
\centering
\begin{tabular}{l r}
\toprule
\textbf{Dataset Name} & \textbf{Count} \\
\midrule
Berkeley Autolab Ur5~\cite{BerkeleyUR5Website} & 892 \\
Bridgev2~\cite{walke2023bridgev2} & 24879 \\
Cmu Play Fusion~\cite{chen2023cmuplayfusion} & 576 \\
Fractal~\cite{brohan2022rt1} & 65530 \\
Kuka~\cite{kalashnikov2018kuka} & 84202 \\
Maniskill~\cite{gu2023maniskill2} & 30029 \\
Taco Play~\cite{rosete2023taco_play} & 3242 \\
Toto~\cite{zhou2023toto} & 899 \\
Utaustin Mutex~\cite{shah2023utaustin_mutex} & 1500 \\
Viola~\cite{zhu2023viola} & 135 \\
Calvin~\cite{mees2022calvin} & 22966 \\
Libero~\cite{liu2023libero} & 1693 \\
\midrule
\textbf{Total} & \textbf{236543} \\
\bottomrule
\end{tabular}
\label{tab:dataset_counts}
\end{table}

\begin{table*}[t]
\centering
\caption{
Long-horizon robotic manipulation evaluation on the CALVIN~\cite{mees2022calvin} benchmark.
Methods marked with ${\dagger}$
 are from our re-implementation.
}
    \label{tab:eval_calvin}
\setlength{\tabcolsep}{5pt}
\begin{tabular}{lccccccc}
\toprule
\multirow{2}{*}{Method} & \multirow{2}{*}{Task} & \multicolumn{5}{c}{Tasks Completed in a Row} & \multirow{2}{*}{\textbf{Avg. Len} $\uparrow$} \\
\cmidrule(lr){3-7}
& & 1 & 2 & 3 & 4 & 5 & \\
\midrule
UniVLA$^{\dagger}$~\cite{wang2025unified} & \multirow{2}{*}{ABCD$\rightarrow$D} & \textbf{0.988} &	0.934 & 0.883 & 0.829 & 0.764 & 4.398  \\
\textbf{Ours} &  &  0.972 & \textbf{0.939} & \textbf{0.894} & \textbf{0.859} & \textbf{0.809} & \textbf{4.473}  \\
\midrule
TLA~\cite{bu2025tla} & \multirow{4}{*}{ABC$\rightarrow$D} & 0.955 & 0.858 & 0.754 & 0.669 & 0.565 & 3.800  \\
Dita~\cite{hou2025dita} &  & 0.945 & 0.825 & 0.728 & 0.613 & 0.500 & 3.610 \\
UniVLA$^{\dagger}$~\cite{wang2025unified} &  &  \textbf{0.972} & 0.902 & 0.826 & 0.741 & 0.661 & 4.102 \\
\textbf{Ours} &  & 0.968 & \textbf{0.912} & \textbf{0.844} & \textbf{0.779} & \textbf{0.708} & \textbf{4.211} \\
\bottomrule
\end{tabular}
\end{table*}

\begin{table*}[t]
    \caption{Evaluation on SimplerEnv-Google Robot~\cite{li2024evaluating} across various manipulation tasks.}
    \label{tab:eval_simpler_google}
    \centering
    \scalebox{1}{
    \begin{tabular}{lccccccc} 
    \toprule
     Model & Pick & Move & Drawer & Place  & \textbf{Average} \\
\midrule
OpenVLA~\cite{kim2024openvla} & 0.180 & 0.563 & 0.630 & 0.000 & 0.343  \\
SpatialVLA~\cite{qu2025spatialvla} & 0.860 & \textbf{0.779} & 0.574 & 0.090 & 0.576  \\
MoTo~\cite{chen2025moto} & 0.740 & 0.604 & 0.431 & 0.000 & 0.444  \\
villa-X~\cite{chen2025villa} & \textbf{0.987} & 0.750 & \textbf{0.593} & 0.056 & 0.597 \\
UniVLA~\cite{wang2025unified} & 0.870 & 0.565 & 0.194 & 0.167 & 0.449   \\
\midrule
\rowcolor{gray!20}
Ours & 0.923 & 0.676 & 0.428 & \textbf{0.407} & \textbf{0.609}  \\
\bottomrule
    \end{tabular}
    }
\end{table*}

\begin{figure}[t]
  \centering
  \includegraphics[width=1\linewidth,
                   trim=0 8 0 2, clip]{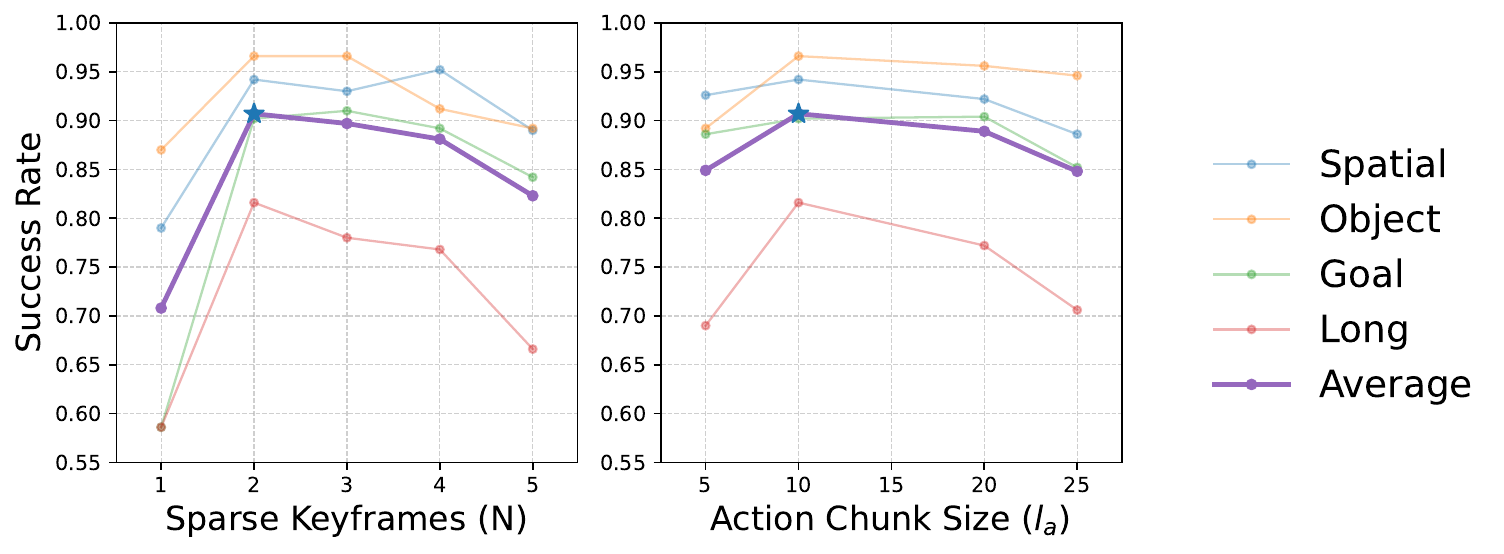}
  \caption{Sensitivity analysis of $N$ and $l_a$ on LIBERO.}
  \label{fig:sensitivity}
\end{figure}

\begin{table*}[t]
\caption{
Comparison between our latent motion representation and Wan~2.1 VAE latent $\mathbf{z}$ on LIBERO.
}
\centering
\label{tab:chain_of_world_ablation}
\resizebox{0.8\linewidth}{!}{
\begin{tabular}{lcc|cccccc}
\toprule
\textbf{Variant} & \textbf{Pre-training} & \textbf{Co-fine-tuning} & Spatial & Object & Goal & Long & \textbf{Average} \\
\midrule
Ours
& latent motion + terminal frame 
& + latent motion 
& 0.948 & 0.974 & 0.958 & 0.906 & \textbf{0.947} \\
\midrule
Wan2.1 VAE~\cite{wan2025wan}
& latent $\mathbf{z}$ + terminal frame 
& + latent $\mathbf{z}$ 
& 0.938 & 0.950 & 0.922 & 0.868 & 0.920 \\
\bottomrule
\end{tabular}
}
\end{table*}

\begin{figure*}[t]
  \centering   
\includegraphics[width=\linewidth]{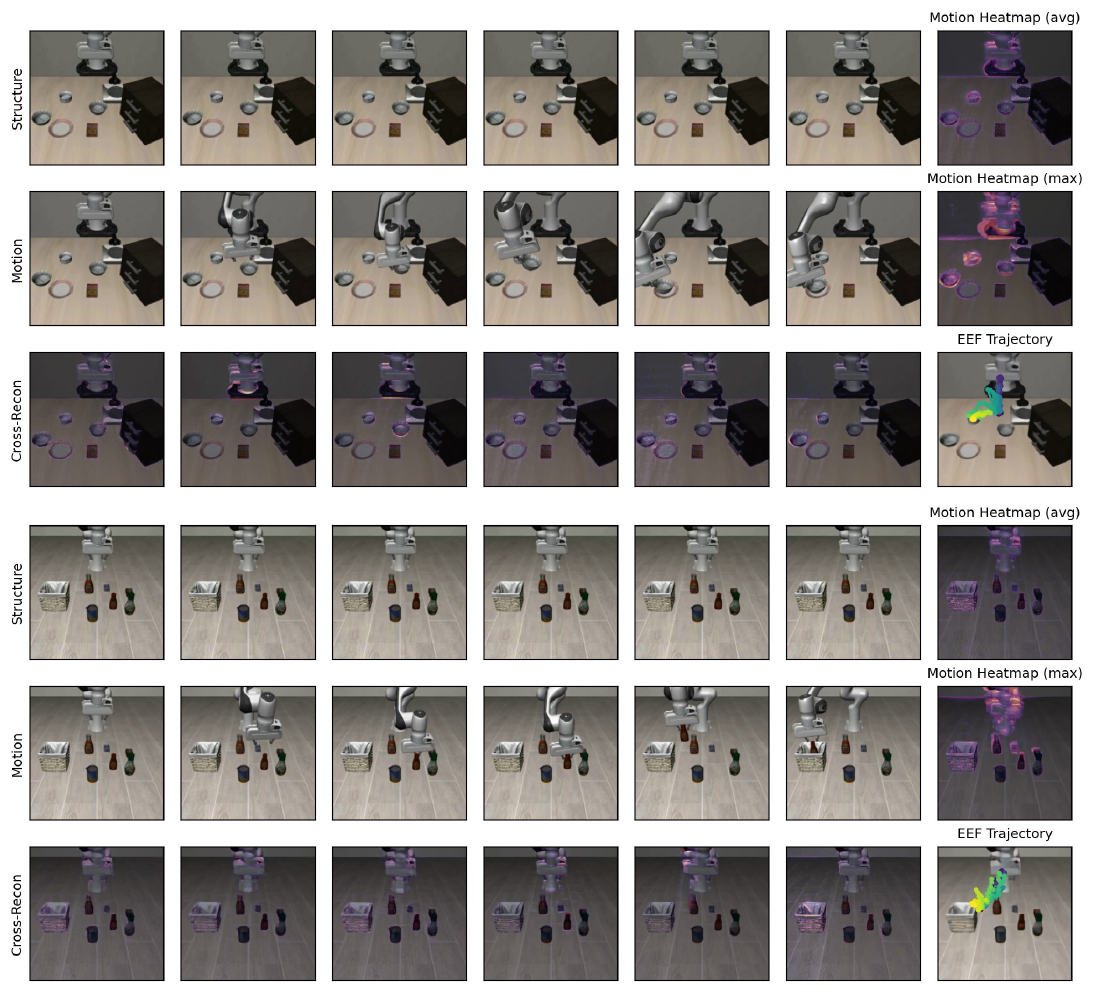}
   \caption{Cross-Recon visualization on LIBERO~\cite{liu2023libero}.
The first six columns show temporally sampled frames from three rows: Structure (top), Motion (middle), and Cross-Recon (bottom).
The Cross-Recon videos are generated by combining the static appearance from the Structure video with the motion representation extracted from the Motion video, revealing the transferred motion patterns.
Each Cross-Recon frame is overlaid with a motion heatmap to highlight dynamic regions.
The last column presents three summary maps: motion heatmaps obtained by averaging and maximizing per-frame absolute differences between Cross-Recon and Structure, and the end-effector trajectory estimated from the motion regions.
}
   \label{fig:motion_vis_libero}
\end{figure*}

\begin{figure*}[t]
  \centering   
\includegraphics[width=\linewidth]{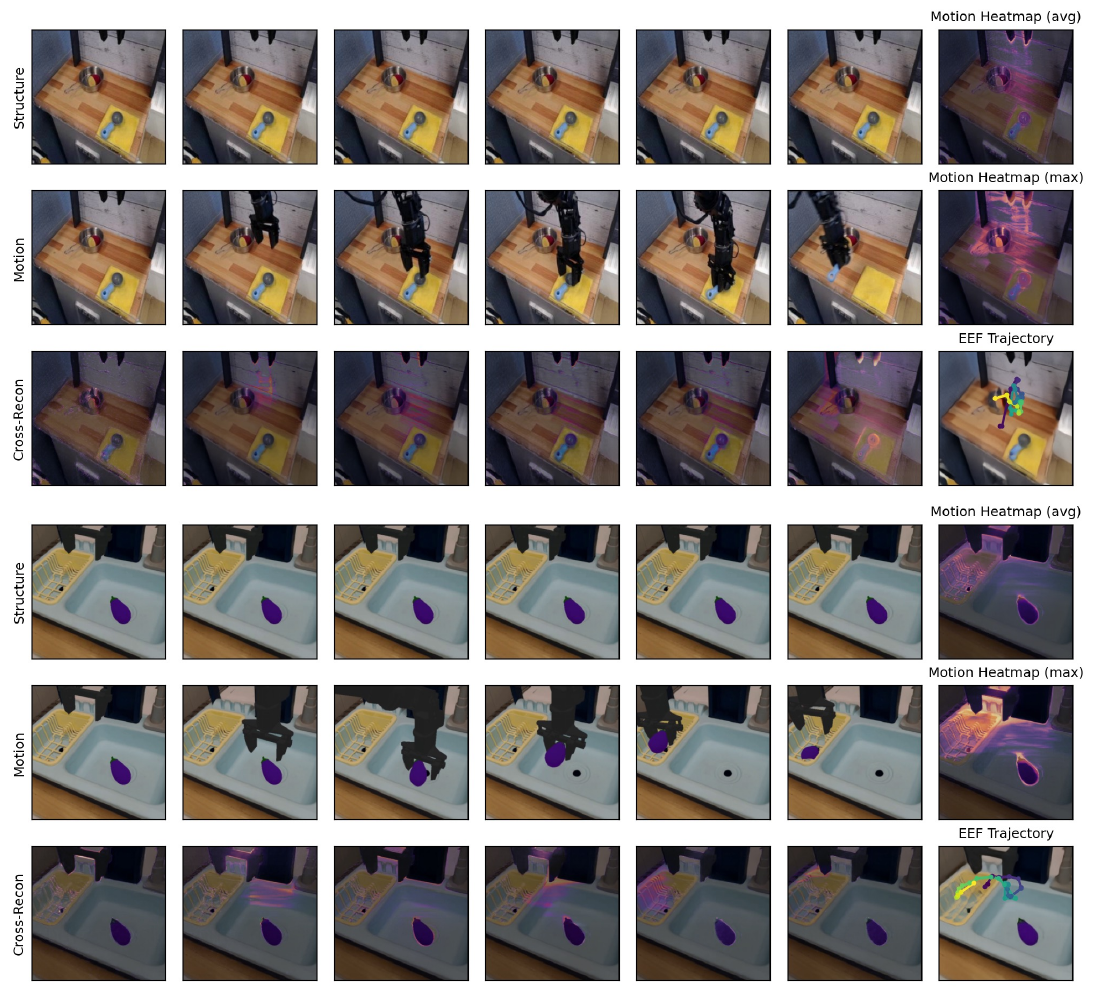}
   \caption{Cross-Recon visualization on SimplerEnv~\cite{li2024evaluating} and Bridgev2~\cite{walke2023bridgev2}.}
   \label{fig:motion_vis_bridge}
\end{figure*}

\begin{figure*}[t]
  \centering   
\includegraphics[width=\linewidth]{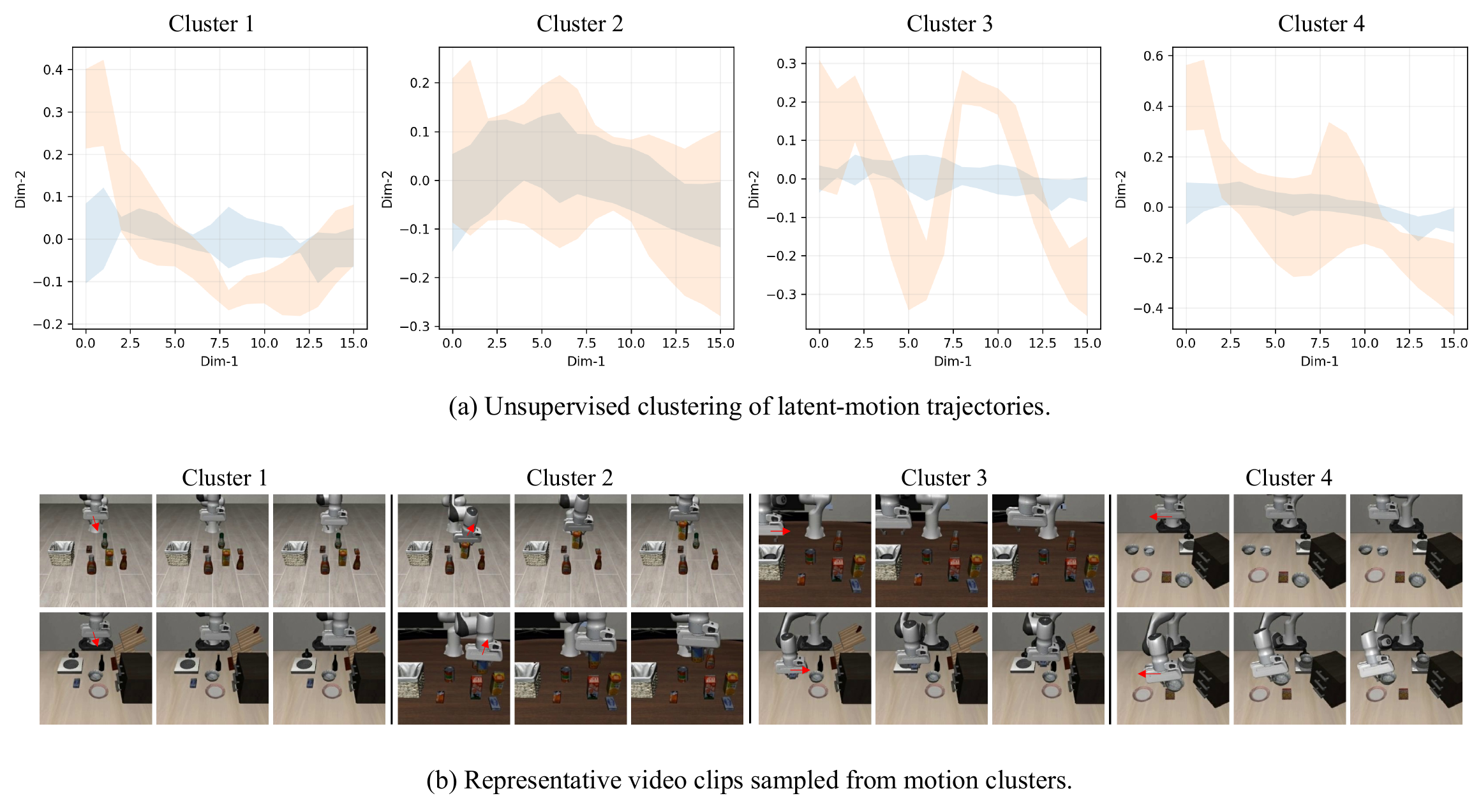}
   \caption{Visualization of latent-motion clusters and corresponding video examples.
(a) Unsupervised clustering results of clip-level motion trajectories. Each subplot shows the average 2D motion trajectory (obtained from the first two PCA components of the accumulated frame-wise motion deltas) for one cluster.
   (b) Representative video examples from clusters.
   Cluster 1 and 2 correspond to monotonic downward-like or upward-like motions, whereas Cluster 3 and 4 exhibit rightward-like or leftward-like behaviors.
   }
   \label{fig:motion_tsne_libero_vis}
\end{figure*}

\begin{figure*}[t]
  \centering   
\includegraphics[width=\linewidth]{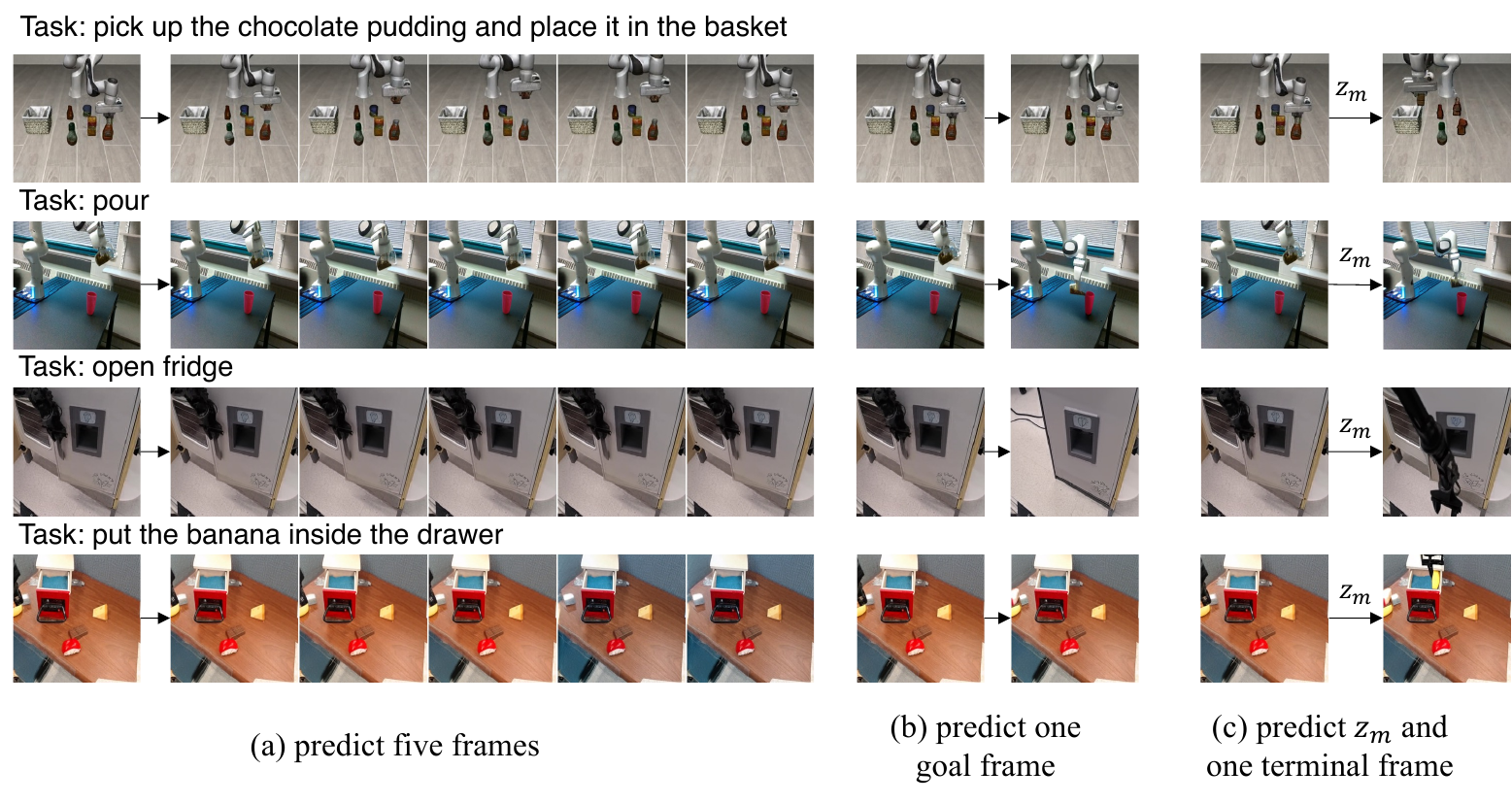}
   \caption{Comparative visualization of future-frame prediction strategies.}
   \label{fig:univla_cotvla_ours_supp}
\end{figure*}

\subsection{Interpretation of the World Model and the Latent Motion Chain}

Our method combines a world model formulation with latent action modeling. 
The world model component consists of two stages: pre-training and co-fine-tuning.
During pre-training, the world model is not action-conditioned. 
This follows the representation adopted by UniVLA~\cite{wang2025unified} and FlowVLA~\cite{zhong2025flowvla}, where the world model predicts future environment evolution given a language instruction and an initial state, rather than explicit actions.
During the co-fine-tuning stage, we introduce an action-conditioned formulation: $p(v^{t+1} \mid v^t, A^t)$.

Our latent motion does not explicitly perform multi-step rollouts. Instead, it provides a continuous and decoupled motion encoding over a temporal window, which can be interpreted as an implicit motion chain.

\section{Additional Results}
\subsection{Analysis of keyframes and action chunk size}

We evaluate the number of sparse keyframes $N\!\in\!\{1,2,3,4,5\}$ and action chunk sizes $l_a\!\in\!\{5,10,20,25\}$ on LIBERO to understand the temporal granularity required by latent motion reasoning.
As shown in Figure~\ref{fig:sensitivity}, both hyperparameters exhibit a clear inverted-U trend. The best performance is achieved at $(N=2, l_a=10)$, corresponding to a $\sim$20-frame ($\approx$2\,s) temporal horizon.

When using only one keyframe ($N=1$), performance drops significantly across all suites, especially on long-horizon tasks, indicating that the latent motion becomes under-constrained. Increasing $N$ to 2 provides sufficient visual anchoring and yields the largest improvement. However, further increasing $N$ gradually degrades performance. With dense observations, the model can rely on short-term visual matching instead of inferring motion dynamics, weakening the benefit of latent temporal reasoning.

A similar phenomenon appears for action chunk size. Small chunks ($l_a=5$) reduce temporal abstraction and make the policy closer to step-wise imitation. Large chunks ($l_a\ge20$) introduce high uncertainty in future evolution, particularly harming the long-horizon tasks. The intermediate chunk size ($l_a=10$) achieves the best trade-off between predictability and abstraction.

Overall, the results suggest that the proposed model performs best when sparse observations provide partial constraints while still requiring the model to infer continuous evolution. This supports our design motivation: the latent motion token serves as a dynamics aggregator over a medium temporal window rather than dense frame tracking or one-step prediction.

\subsection{Comparison with other Video VAE}
To further analyze the role of latent motion representations, we replace VidTwin with the VAE from Wan~2.1~\cite{wan2025wan} and conduct a controlled comparison. Specifically, we use the latent $\mathbf{z}$ extracted by the Wan~2.1 VAE as auxiliary supervision during both pre-training and co-fine-tuning.

The Wan~2.1 VAE is trained on large-scale video data and therefore incorporates rich generic video priors. As shown in Table~\ref{tab:chain_of_world_ablation}, this variant achieves an average success rate of 0.920 on LIBERO. While competitive, it remains inferior to our latent motion design (0.947).

\subsection{CALVIN}
Calvin~\cite{mees2022calvin} is an open-source simulated benchmark built on PyBullet, designed for learning long-horizon, language-conditioned robotic manipulation tasks.
It provides a tabletop simulation environment containing 23 types of manipulation skills, such as lifting, pushing, rotating, and object relocation.
These skills must be executed in sequence to complete multi-step tasks, introducing substantial uncertainty and randomness, which makes Calvin a highly challenging evaluation benchmark. 
The dataset includes a large number of expert demonstrations and is organized into multiple subsets. 
In our experiments, we use the ABCD$\rightarrow$D and ABC$\rightarrow$D subsets, and during training, we only utilize demonstrations that include natural language descriptions of the actions. 
Following the official evaluation protocol, all tests consist of 1000 episodes, each containing a sequence of five sub-tasks specified by natural language instructions.

The main results are presented in Table~\ref{tab:eval_calvin}. Our method achieves an average success length of 4.473 on the ABCD$\rightarrow$D task and 4.211 on the ABC$\rightarrow$D task. 
For a fair comparison, we reproduced UniVLA~\cite{wang2025unified} using the training sets listed in Table~\ref{tab:dataset_counts}, and followed a fine-tuning setup with 16 A800 GPUs and a per-GPU batch size of 8.
Under the same training configuration, our approach outperforms UniVLA~\cite{wang2025unified}.

\subsection{SimplerEnv-Google Robot}
We also evaluate our method on the SimplerEnv-Google Robot benchmark.
The evaluation primarily follows the visual matching protocol, which assesses the alignment between real and simulated visual appearances by overlaying real-world images onto simulated backgrounds and adjusting the textures of foreground objects and the robot within the simulator. 
This benchmark includes four tasks: \textit{pick coke can}, \textit{move near}, \textit{open/close drawer}, and \textit{place in closed drawer}.

The main results are shown in Table~\ref{tab:eval_simpler_google}.
Our method achieves an average success rate of 0.609, outperforming UniVLA~\cite{wang2025unified}, villa-x~\cite{chen2025villa}, MoTo~\cite{chen2025moto}, and other baselines. 
Here, UniVLA refers to our reproduction. 
Our method surpasses UniVLA on all four tasks and shows a particularly large improvement on the \textit{place in closed drawer} task.

\begin{figure}[t]
  \centering   
\includegraphics[width=0.8\linewidth]{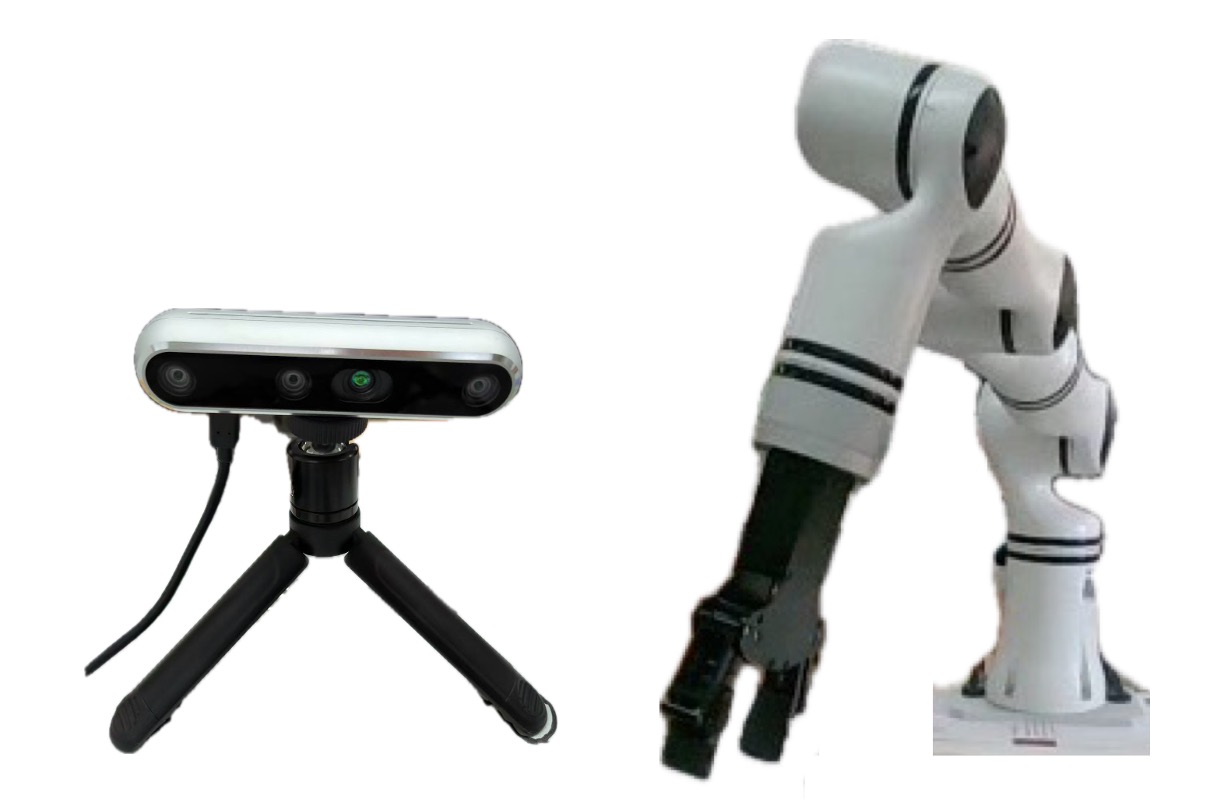}
   \caption{An Intel RealSense camera and a Realman RM75B robot.}
   \label{fig:robot_camera}
\end{figure}

\begin{figure*}[t]
  \centering   
\includegraphics[width=0.9\linewidth]{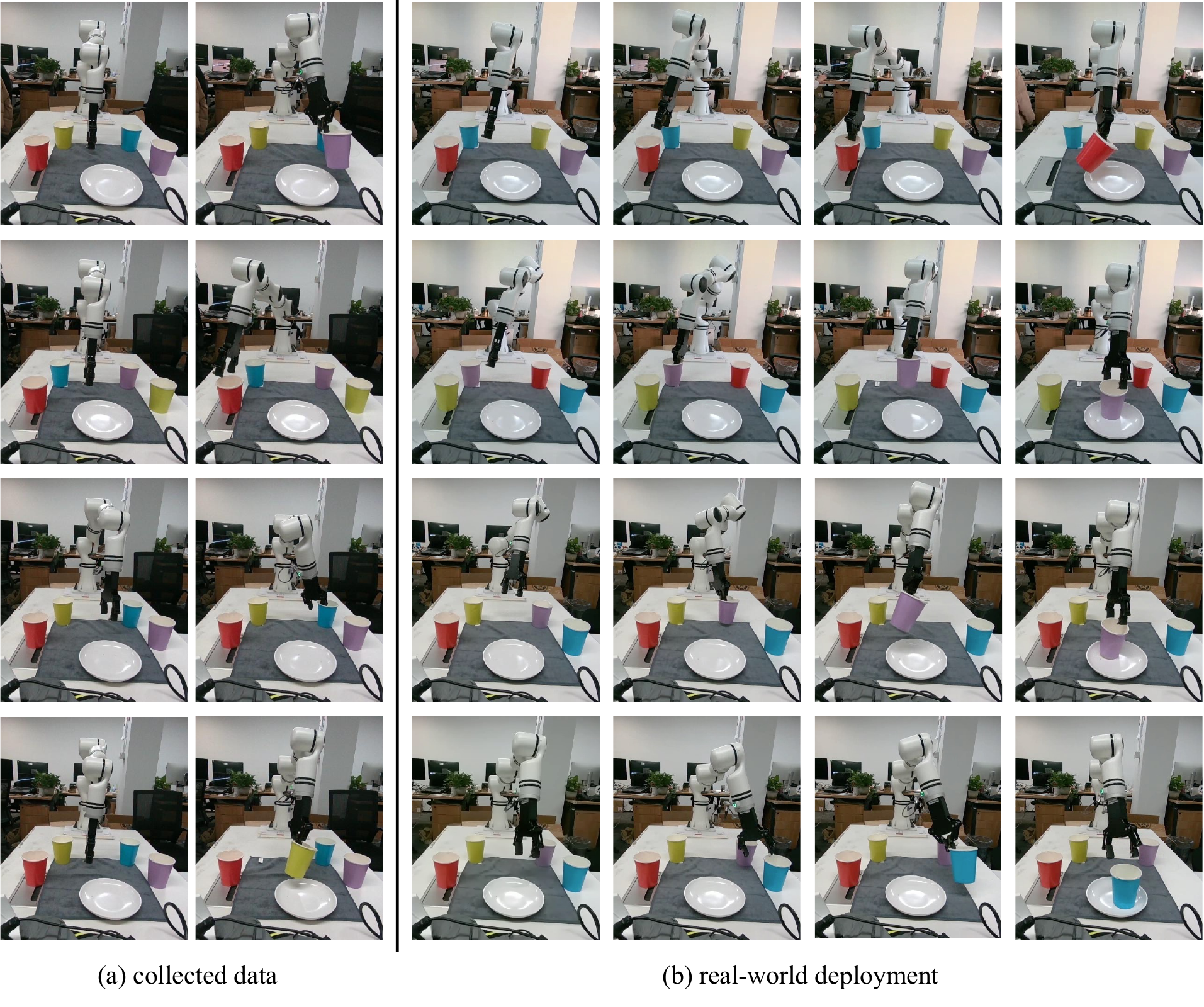}
   \caption{Comparison between data collection and real-world deployment during testing.}
   \label{fig:real_robot}
\end{figure*}

\subsection{More Visualization}
We provide extended visualizations for the latent motion analysis presented in Section~4.4, with the main results shown in Figures~\ref{fig:motion_vis_libero}, \ref{fig:motion_vis_bridge}, \ref{fig:motion_tsne_libero_vis}, and \ref{fig:univla_cotvla_ours_supp}.

\textbf{Effective decoupling of structure and motion latents.}
Figures~\ref{fig:motion_vis_libero} and \ref{fig:motion_vis_bridge} analyze representative samples from the Libero and Bridge datasets. The first six columns display temporally sampled frames from three rows: Structure (top), Motion (middle), and Cross-Reconstruction (bottom). The Cross-Recon videos are synthesized by combining the static appearance from the Structure video with the motion representation extracted from the Motion video, thereby revealing transferred motion patterns. 
Each Cross-Recon frame is overlaid with a motion heatmap to highlight dynamic regions.
The final column summarizes three diagnostic maps: motion heatmaps computed by averaging and maximizing the per-frame absolute differences between Cross-Recon and Structure, as well as the end-effector trajectory estimated from the activated motion regions. As shown, the highlighted areas consistently follow the movement of the robot arm in the Motion video. 
In the video results, these regions fluctuate over time; for clarity in static visualization, we display aggregated highlights in the figures.

We further analyze the distribution of motion latents, as shown in Figure~\ref{fig:motion_tsne_libero_vis}. To derive an interpretable trajectory representation from high-dimensional motion latents, we first extract per-frame motion features from each video clip and accumulate framewise differences to obtain a temporal sequence describing the overall motion trend of the clip. These sequences are then resampled to a fixed length across all clips and standardized globally. We subsequently apply PCA to the sequence features and take the first two principal components as a 2D trajectory for each clip. This representation preserves the dynamic structure encoded in the latent space while enabling clear comparison across clips.

Figure~\ref{fig:motion_tsne_libero_vis} (a) shows unsupervised clustering of all motion trajectories in the 2D PCA space. To obtain cluster-level canonical shapes, we temporally align trajectories within each cluster via resampling and plot their mean curves along with 95\% confidence intervals. Distinct trajectory patterns emerge across clusters—such as monotonic rises, two-stage reversals, and multi-phase back-and-forth motions—indicating that the model’s motion latent captures high-level motion semantics.
To further validate the semantic consistency within each cluster, we randomly sample two video clips per cluster and visualize three uniformly sampled frames from each clip, as shown in Figure~\ref {fig:motion_tsne_libero_vis} (b). The clips within the same cluster exhibit highly similar motion trends in appearance, confirming that the structure of the motion-latent space yields meaningful discrimination among different action patterns.

\textbf{Motion latent enhances dynamic modeling for future frame prediction.}
As shown in Figure~\ref{fig:univla_cotvla_ours_supp}, we further visualize future frame predictions under different pretraining strategies. From top to bottom, the examples correspond to four tasks:
i) pick up the chocolate pudding and place it in the basket,
ii) pour,
iii) open the fridge, and
iv) put the banana inside the drawer.
In Figure~\ref{fig:univla_cotvla_ours_supp} (a), world-model-based approaches suffer from reconstructing redundant background pixels, which can draw attention away from critical interactions and motion cues. As a result, the predicted future frames sometimes remain nearly unchanged, such as in tasks (ii) and (iii).
Figure~\ref{fig:univla_cotvla_ours_supp} (b) shows that predicting only the target frame often leads to unstable generation due to the absence of intermediate evolution steps: in task (i), the target frame nearly collapses back to the initial frame, and in task (iii), only one door of the fridge is generated.
In contrast, our method leverages the motion latent $z_m$ as a chain-of-thought for motion, providing stronger guidance for future-frame prediction. The generated final frames align more accurately with the intended task instructions.

\section{Real-Robot Experiments}

\textbf{Experimental Setup.} 
As shown in Figure~\ref{fig:robot_camera}, we use the Realman RM75B robot, which is equipped with 7 degrees of freedom and a single gripper. 
An Intel RealSense camera is used to capture RGB images.
We set up a cup-grasping experiment and collected a total of 127 episodes, consisting of 65,382 frames with corresponding actions. 
Each episode contains an average of 515 frames, corresponding to approximately 20 seconds in the real world. 
The dataset mainly includes grasping cups of four different colors, with the number of episodes per color as follows: red 31, blue 39, yellow 24, and purple 33.
Figure~\ref{fig:real_robot} (a) shows some collected data.

During training, all images are cropped and resized to 256×256. The action chunk size is set to 10. We train the model for 2k steps using 16 GPUs with a per-GPU batch size of 8.
The data were collected in the afternoon and evening and then used for model training. 
Testing was conducted the following day. 
As shown in Figure~\ref{fig:real_robot}, the lighting conditions have some differences between data collection compared and during real-world deployment. 
We found that the model was still able to correctly execute instructions under different lighting conditions. 
Figure~\ref{fig:real_robot} (b) shows in the first two rows two test cases: grasping a red/purple cup and placing it on a plate. Their background lighting differs from the training data, but the model is still able to execute the tasks successfully.

\end{document}